\documentclass[10pt,twocolumn,letterpaper]{article}

\usepackage{cvpr}
\usepackage{times}
\usepackage{epsfig}
\usepackage{graphicx}
\usepackage{amsmath}
\usepackage{amssymb}
\usepackage{subfigure}
\usepackage{multirow}
\usepackage{booktabs}
\usepackage{array}
\usepackage{color}
\usepackage{caption}
\usepackage{stfloats}
\usepackage{cite}
\usepackage{algorithm}
\usepackage{algorithmic}
\usepackage{url}
\usepackage{overpic}
\usepackage{xspace}
\usepackage{bm}
\usepackage{rotating}
\usepackage{mathptmx}
\usepackage{xfrac}
\usepackage{array}

\usepackage[pagebackref=true,breaklinks=true,letterpaper=true,colorlinks,bookmarks=false]{hyperref}

\cvprfinalcopy 

\def\cvprPaperID{1935} 

\ifcvprfinal\fi

\newcommand{\PreserveBackslash}[1]{\let\temp=\\#1\let\\=\temp}
\newcolumntype{C}[1]{>{\PreserveBackslash\centering}p{#1}}
\newcolumntype{R}[1]{>{\PreserveBackslash\raggedleft}p{#1}}
\newcolumntype{L}[1]{>{\PreserveBackslash\raggedright}p{#1}}
\newcommand{\tabincell}[2]{\begin{tabular}{@{}#1@{}}#2\end{tabular}}
\newcommand\blfootnote[1]{%
  \begingroup
  \renewcommand\thefootnote{}\footnote{#1}%
  \addtocounter{footnote}{-1}%
  \endgroup
}
\begin{document}

\title{GA-Net: Guided Aggregation Net for End-to-end Stereo Matching}

\author{Feihu Zhang$^{1 *}$\qquad Victor Prisacariu$^1$\qquad Ruigang Yang$^2$ \qquad Philip H.S. Torr$^1$\\
	$^1$ University of Oxford\quad\qquad\qquad\qquad\qquad $^2$ Baidu Research, Baidu Inc.
}

\maketitle

\begin{abstract}
 In the stereo matching task, matching cost aggregation is crucial in both traditional methods and deep neural network models in order to accurately estimate disparities. 
 We propose two novel neural net layers, aimed at capturing local and the whole-image cost dependencies respectively.  The first is a semi-global aggregation layer which is a differentiable approximation of the semi-global matching, the second is the local guided aggregation layer which follows a traditional cost filtering
 strategy to refine thin structures. 
 
 These two layers can be used to replace the widely used 3D convolutional layer which is computationally costly and memory-consuming as it has cubic computational/memory complexity. In the experiments, we show that nets with a two-layer guided aggregation block easily outperform the state-of-the-art GC-Net which has nineteen 3D convolutional layers. We also train a deep guided aggregation network (GA-Net) which gets better accuracies than state-of-the-art methods on both Scene Flow dataset and KITTI benchmarks. {\color{magenta}{\small  Code will be available at https://github.com/feihuzhang/GANet.}}

\end{abstract}

\vspace{-1mm}
\section{Introduction}
\vspace{-1mm}
Stereo reconstruction is a major research topic in computer vision, robotics and autonomous driving. It aims to estimate 3D geometry by computing disparities between matching pixels in a stereo image pair. It is challenging due to a variety of real-world problems, such as occlusions, large textureless areas (\eg sky, walls \etc), reflective surfaces (\eg windows), thin structures and repetitive textures.
\blfootnote{\hspace{-2mm}$^*$Part of the work was done when working in Baidu Research.}
\vspace{-4mm}

\begin{figure}[t]
\setlength{\abovecaptionskip}{0pt}
\setlength{\belowcaptionskip}{-10pt}
\vspace{-1mm}
\centering
\subfigure[Input image]{
\includegraphics[width=0.475\linewidth,height=0.29\linewidth]{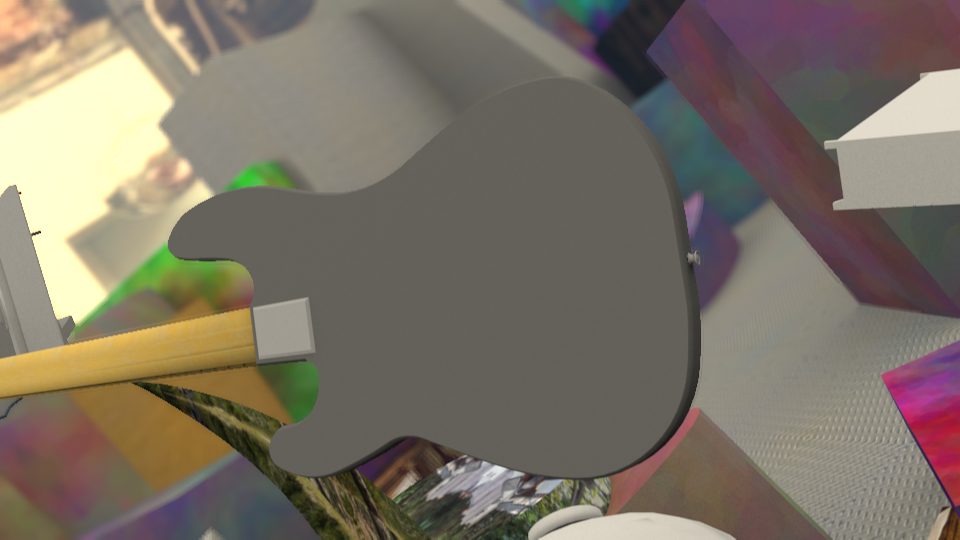}
}
\hspace{-2mm}
\subfigure[GC-Net\cite{kendall2017end}]{
\includegraphics[width=0.475\linewidth,height=0.29\linewidth]{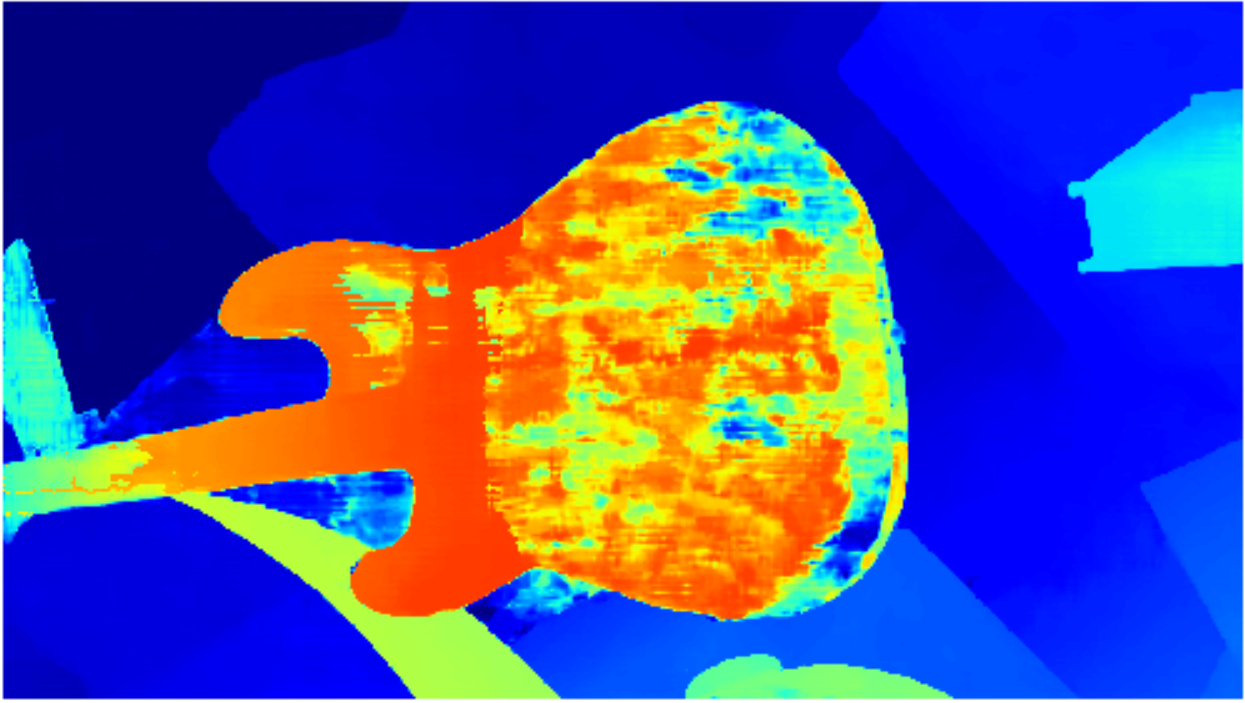}
}
\\[-3mm]
\subfigure[Our GA-Net-2]{
\includegraphics[width=0.475\linewidth,height=0.29\linewidth]{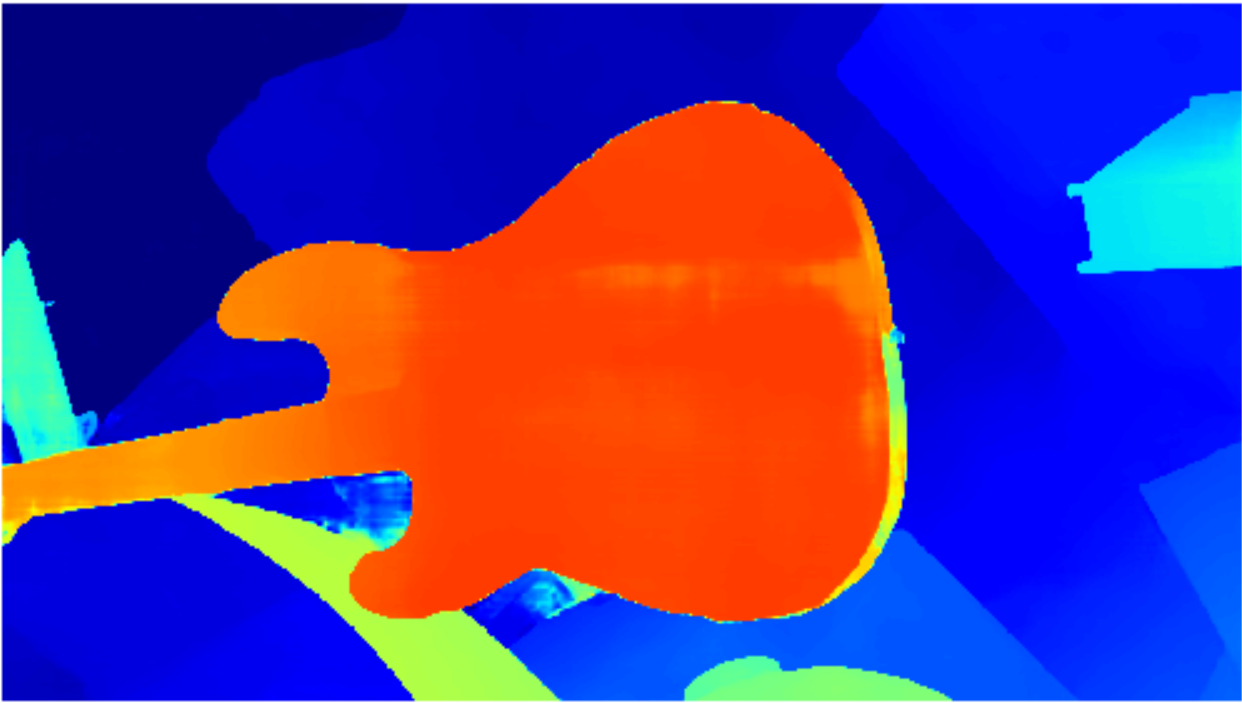}
}
\hspace{-2mm}
\subfigure[Ground truth]{
\includegraphics[width=0.475\linewidth,height=0.29\linewidth]{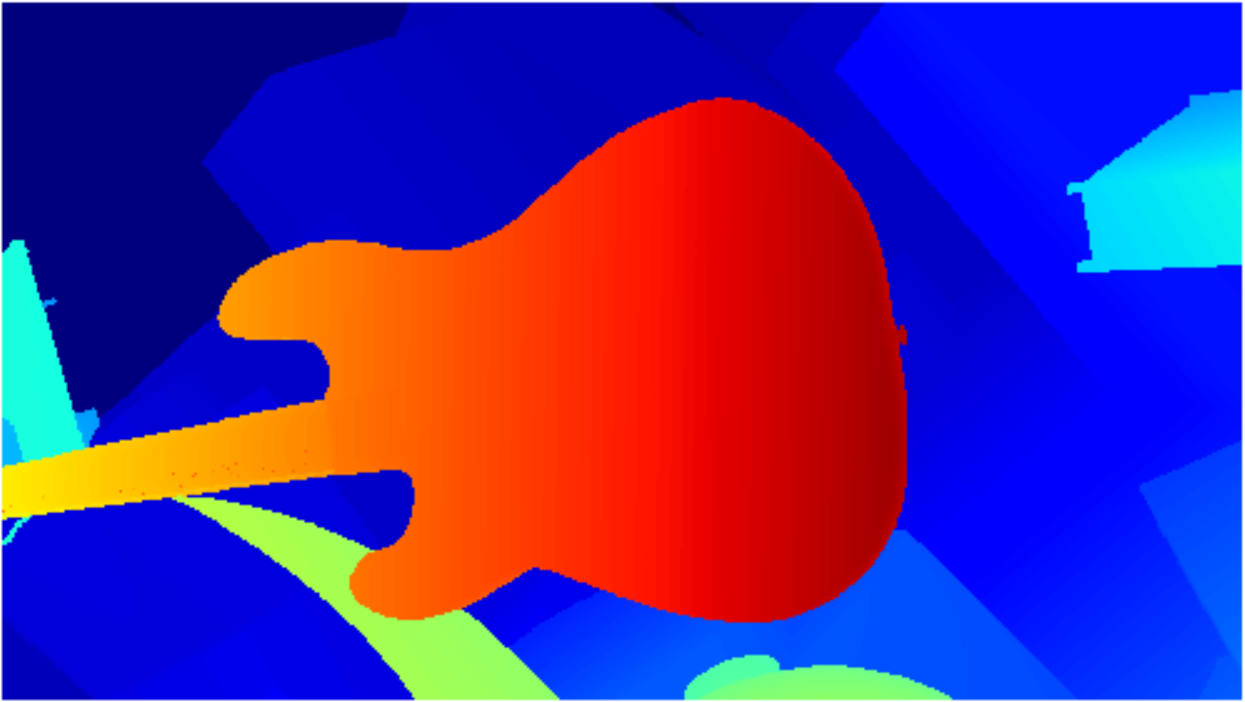}
}
\caption{\small Performance illustrations. (a) a challenging input image.  (b) Result of the state-of-the-art method GC-Net \cite{kendall2017end} which has nineteen 3D convolutional layers for matching cost aggregation. (c) Result of our GA-Net-2, which only uses two proposed GA layers and two 3D convolutional layers. It aggregates the matching information into the large textureless region and is an order of magnitude faster than GC-Net. (d) Ground truth.}
\label{fig:illustration}
\end{figure}
Traditionally, stereo reconstruction is decomposed into three important steps:  feature extraction (for matching cost computation),  matching cost aggregation and disparity prediction \cite{scharstein2002taxonomy,hirschmuller2008stereo}. Feature-based matching is often ambiguous, with wrong matches having a lower cost than the correct ones, due to occlusions, smoothness, reflections, noise etc. Therefore, cost aggregation is a key step needed to obtain accurate disparity estimations in challenging regions.

Deep neural networks have been used for matching cost computation in, e.g, \cite{zbontar2015computing, zhang2018fundamental}, with (i) cost aggregation based on traditional approaches, such as cost filtering \cite{hosni2013fast} and semi-global matching (SGM) \cite{hirschmuller2008stereo} and (ii) disparity computation with a separate step. Such methods considerably improve over traditional pixel matching, but still struggle to produce accurate disparity results in textureless, reflective and occluded regions. End-to-end approaches that link matching with disparity estimation were developed in e.g. DispNet \cite{mayer2016large}, but it was not until GC-Net \cite{kendall2017end} that cost aggregation, through the use of 3D convolutions, was incorporated in the training pipeline. The more recent work of \cite{chang2018pyramid}, PSMNet, further improves accuracy by implementing the stacked hourglass backbone \cite{newell2016stacked} and considerably increasing the number of 3D convolutional layers for cost aggregation. The large memory and computation cost incurred by using 3D convolutions is reduced by down-sampling and up-sampling frequently, but this leads to a loss of precision in the disparity map.

Among these approaches, traditional semi-global matching (SGM) \cite{hirschmuller2008stereo} and  cost filtering \cite{hosni2013fast} are all robust and efficient cost aggregation methods which have been widely used in many industrial products. But, they are not differentiable  and cannot be easily trained in an end-to-end manner.

In this work, we propose two novel cost aggregation layers for end-to-end stereo reconstruction to replace the use of 3D convolutions. Our solution considerably increases accuracy, while decreasing both memory and computation costs. 

First, we introduce a semi-global guided aggregation layer (SGA) which implements a differentiable approximation of semi-global matching (SGM) \cite{hirschmuller2008stereo} and aggregates the matching cost in different directions over the whole image. This enables accurate estimations in occluded regions or large textureless/reflective regions. 

Second, we introduce a local guided aggregation  layer (LGA) to cope with thin structures and object edges in order to recover the loss of details caused by down-sampling and up-sampling layers.

As illustrated in Fig. \ref{fig:illustration}, a cost aggregation block with only two GA layers and two 3D convolutional layers easily outperforms the state-of-the-art GC-Net \cite{kendall2017end}, which has nineteen 3D convolutional layers. More importantly, one GA layer has only 1/100 computational complexity in terms of FLOPs (floating-point operations) as that of a 3D convolution. This allows us to build a real-time GA-Net model, which achieves better accuracy compared with other existing real-time algorithms and runs at a speed of 15$\sim$20 fps.

We further increase the accuracy by 
improving the network architectures used for feature extraction and matching cost aggregation. 
The full model, which we call ``GA-Net'', achieves the state-of-the-art accuracy on both the Scene Flow dataset \cite{mayer2016large} and the KITTI benchmarks \cite{kitti2012,kitti2015}.

\section{Related Work}
Feature based matching cost is often ambiguous, as wrong matches can easily have a lower cost than correct ones, due to occlusions, smoothness, reflections, noise etc. To deal with this, many cost aggregation approaches have been developed to refine the cost volume and achieve better estimations.
This section briefly introduces related work in the application of deep neural networks in stereo reconstruction with a focus on the existing matching cost aggregation strategies, and briefly reviews approaches for traditional local and semi-global cost aggregations.
\subsection{Deep Neural Networks for Stereo Matching}
Deep neural networks were used to compute patch-wise similarity scores in \cite{zagoruyko2015learning,zhang2018fundamental,chen2015deep,flynn2016deepstereo}, with traditional cost aggregation and disparity computation/refinement methods \cite{hirschmuller2008stereo,hosni2013fast} used to get the final disparity maps. These approaches achieved state-of-the-art accuracy, but, limited by the traditional matching cost aggregation step, often produced wrong predictions in occluded regions, large textureless/reflective regions and around object edges. Some other methods looked to improve the performance of traditional cost aggregation, with, e.g. SGM-Nets \cite{SGMNet} predicting the penalty-parameters for SGM \cite{hirschmuller2008stereo} using a neural net, whereas Sch{\"o}nberger \etal \cite{schonberger2018learning} learned to fuse proposals by optimization in stereo matching and Yang \etal proposed to aggregate costs using a minimum spanning tree \cite{yang2012non}.

Recently, end-to-end deep neural network models have become popular. Mayer \etal created a large synthetic dataset to train end-to-end deep neural network for disparity estimation (\eg DispNet) \cite{mayer2016large}. Pang \etal \cite{Pang_2017} built a two-stage convolutional neural network to first estimate and then refine the disparity maps. Tulyakov \etal proposed end-to-end deep stereo models for practical applications \cite{tulyakov2018practical}. GC-Net \cite{kendall2017end} incorporated the feature extraction, matching cost aggregation and disparity estimation into a single end-to-end deep neural model to get state-of-the-art accuracy on several benchmarks. PSMNet\cite{chang2018pyramid} used pyramid feature extraction and a stacked hourglass block \cite{Newell_2016} with twenty-five 3D convolutional layers to further improve the accuracy.
\subsection{Cost Aggregation}
Traditional stereo matching algorithms \cite{hirschmuller2008stereo,PMPM,PMBP} added an additional constraint to enforce smoothness by penalizing changes of neighboring disparities. This can be both local and (semi-)global, as described below.
\vspace{-3mm}
\subsubsection{Local Cost Aggregation}
The cost volume $C$ is formed of matching costs at each pixel's location for each candidate disparity value $d$. It has a size of $H\times W \times D_{max}$ (with $H$: image height, $W$: image width, $D_{max}$: maximum of the disparities) and can be sliced into $D_{max}$ slices for each candidate disparity $d$. 
An efficient cost aggregation method is the local cost filter framework \cite{hosni2013fast,zhang2015segment}, where each slice of the cost volume $C(d)$ is filtered independently by a guided image filter \cite{he2013guided,tomasi1998bilateral,zhang2015segment}. The filtering for pixel's location $\mathbf{p}=(x,y)$ at disparity $d$ is a weighted average of all neighborhoods $\mathbf{q}\in N_\mathbf{p}$ in the same slice $C(d)$:
\begin{equation}
\setlength{\abovedisplayskip}{4pt}
\setlength{\belowdisplayskip}{4pt}
C^A(\mathbf{p},d)=\sum_{\mathbf{q}\in N_\mathbf{p}}{\omega({\mathbf{p},\mathbf{q}})\cdot C(\mathbf{q},d)}
\label{EQ:costfilter}
\end{equation}
Where $C(\mathbf{q},d)$ means the matching cost at location $\mathbf{p}$ for candidate disparity $d$. $C^A(\mathbf{p},d)$ represents the aggregated matching cost.
Different image filters \cite{he2013guided,tomasi1998bilateral,zhang2015segment} can be used to produce the guided filter weights $\omega$. Since these methods only aggregate the cost in a local region $N_\mathbf{p}$, they can run at fast speeds and reach real-time performance.
\vspace{-3.5mm}
\subsubsection{Semi-Global Matching}
\vspace{-0.5mm}
When enforcing (semi-)global aggregation, the matching cost and the smoothness constraints are formulated into one energy function $E(D)$ \cite{hirschmuller2008stereo} with the disparity map of the input image as $D$.  The problem of stereo matching can now be formulated as finding the best disparity map $D^*$ that minimizes the energy $E(D)$: 
\begin{equation}
 \setlength{\abovedisplayskip}{5pt}
\setlength{\belowdisplayskip}{4pt}
\begin{array}{ll}
{\hspace{-4mm}}E(D)=\sum_{\mathbf{p}}\{C_\mathbf{p}(D_\mathbf{p}) & \hspace{-3mm} +\hspace{0.5mm} \sum_{\mathbf{q}\in N_\mathbf{p}}{P_1\cdot\delta(|D_\mathbf{p}-D_\mathbf{q}|=1)}\\[3mm]
 & \hspace{-3mm}+\hspace{0.5mm} \sum_{\mathbf{q}\in N_\mathbf{p}}{P_2\cdot\delta(|D_\mathbf{p}-D_\mathbf{q}|>1)}\}.
\end{array}
\label{EQ:optimize}
\end{equation}
The first term $\sum_{\mathbf{p}}{C_\mathbf{p}(D_\mathbf{p})}$ is the sum of matching costs at all pixel locations $\mathbf{p}$ for  disparity map $D$. The second term is a constant penalty $P_1$ for locations $\mathbf{q}$ in the neighborhood of $\mathbf{p}$ if they have small disparity  discontinuities in disparity map $D$ ($|D_\mathbf{p}-D_\mathbf{q}|=1$). The last term adds a larger constant penalty $P_2$, for all larger disparity changes ($|D_\mathbf{p}-D_\mathbf{q}|>1$).

Hirschmuller proposed to aggregate matching costs in 1D from sixteen directions to get a approximate solution with $O(KN)$ time complexity, which is well known as semi-global matching (SGM) \cite{hirschmuller2008stereo}.
The cost $C^A_\mathbf{r}(\mathbf{p},d)$ of a location $\mathbf{p}$ at disparity $d$ aggregates along a path over the whole image in the direction $\mathbf{r}$, and is defined recursively as:
\begin{equation}
 \setlength{\abovedisplayskip}{3pt}
\setlength{\belowdisplayskip}{3pt}
C^A_\mathbf{r}(\mathbf{p},d)=C(\mathbf{p},d)+\min\left\{\begin{array}{l}
\hspace{-2mm} C^A_\mathbf{r}(\mathbf{p}-\mathbf{r},d),\\
\hspace{-2mm} C^A_\mathbf{r}(\mathbf{p}-\mathbf{r},d-1)+P_1,\\
\hspace{-2mm} C^A_\mathbf{r}(\mathbf{p}-\mathbf{r},d+1)+P_1,\\
\hspace{-2mm} \min\limits_{i}C^A_\mathbf{r}(\mathbf{p}-\mathbf{r},i)+P_2.
\end{array}\right.
\label{EQ:sgm_agg}
\end{equation}
Where $\mathbf{r}$ is a unit direction vector. The same aggregation steps were used in MC-CNN \cite{zbontar2015computing,SGMNet}, and similar iterative steps were employed in \cite{PMBP,bleyer2011patchmatch,liu2017learning}.

In the following section, we detail our much more efficient guided aggregation (GA) strategies, which include a semi-global aggregation (SGA) layer and a local guided aggregation (LGA) layer. Both GA layers can be implemented with back propagation in end-to-end models to replace the low-efficient 3D convolutions and obtain higher accuracy.

\section{Guided Aggregation Net}
In this section, we describe our proposed guided aggregation network (GA-Net), including the guided aggregation (GA) layers and the improved network architecture.

\begin{figure*}[t]
\setlength{\abovecaptionskip}{14pt}
\setlength{\belowcaptionskip}{-8pt}
\vspace{-2.25mm}
\centering
\hspace{8mm}
	\begin{overpic}[width=0.92\linewidth]{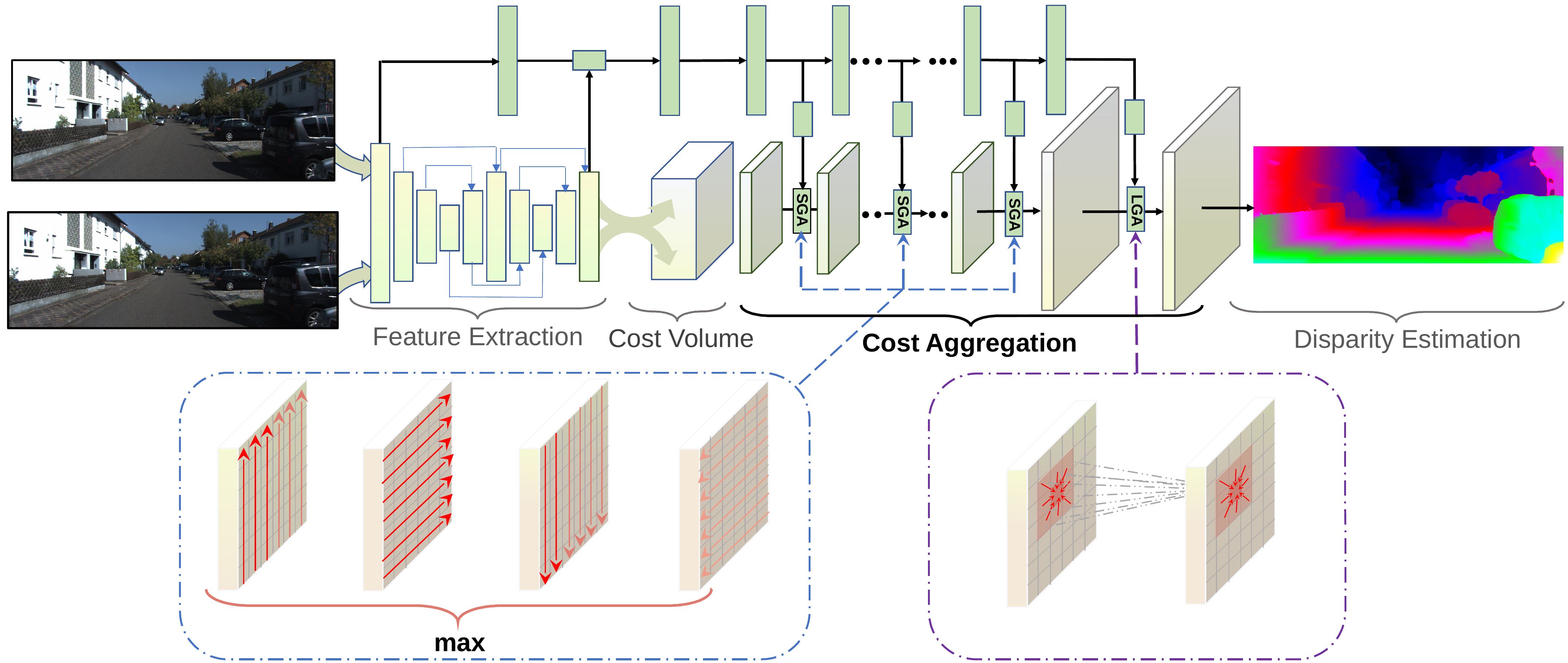}
        \put(-2,40){\small \color{black}{\begin{turn}{270}(a) GA-Net Architecture \end{turn}}}
        \put(14.35,-1.7){\small \color{black}{(b) Semi-Global Guided Aggregation (SGA)}}
         \put(58.2,-1.7){\small \color{black}{(c) Local Guided Aggregation (LGA)}}
        \end{overpic}
\caption{\small (a) Architecture overview. The left and right images are fed to a weight-sharing feature extraction pipeline. It
consists of a stacked hourglass CNN and is connected by concatenations.  The  extracted left and right image features are then used to form a 4D cost volume, which is fed into a cost aggregation block for regularization, refinement and disparity regression. The guidance subnet (green) generates the weight matrices for the guided cost aggregations (SGA and LGA). (b) SGA layers semi-globally aggregate the cost volume in four directions. (c) The LGA layer is used before the disparity regression and locally refines the 4D cost volume for several times.}
\label{fig:architecture}
\end{figure*}

\subsection{Guided Aggregation Layers}
State-of-the-art end-to-end stereo matching neural nets such as \cite{kendall2017end,chang2018pyramid} build a 4D matching cost volume (with size of $H\times W\times D_{max}\times F$, $H$: height, $W$: width, $D_{max}$: max disparity, $F$: feature size) by concatenating features between the stereo views, computed at different disparity values. This is next refined by a cost aggregation stage, and finally used for disparity estimation. Different from these approaches, and inspired by semi-global and local matching cost aggregation methods \cite{hirschmuller2008stereo,hosni2013fast}, we propose our semi-global guided aggregation (SGA) and local guided aggregation (LGA) layers, as outlined below.

\subsubsection{Semi-Global Aggregation}

Traditional SGM \cite{hirschmuller2008stereo} aggregates the matching cost iteratively in different directions (Eq. (\ref{EQ:sgm_agg})). There are several difficulties in using such a method in end-to-end trainable deep neural network models.

First, SGM has many user-defined parameters ($P_1, P_2$), which are not straightforward to tune. All of these parameters become unstable factors during neural network training. Second, the cost aggregations and penalties in SGM are fixed for all pixels, regions and images without adaptation to different conditions.
Third, the hard-minimum selection leads to a lot of fronto parallel surfaces in depth estimations.

We design a new semi-global cost aggregation step which supports backpropagation. This is more effective than the traditional SGM and can be used repetitively in a deep neural network model to boost the cost aggregation effects. The proposed aggregation step is:
\begin{equation}
\setlength{\abovedisplayskip}{4pt}
\setlength{\belowdisplayskip}{4pt}
\begin{array}{rll}
C^A_\mathbf{r}(\mathbf{p},d)&\hspace{-2mm}=&\hspace{-2mm}C(\mathbf{p},d)\\
&\hspace{-8mm}+&\hspace{-8mm}\text{sum}\left\{\begin{array}{l}
\hspace{-2mm} {\mathbf{w}_1(\mathbf{p},\mathbf{r})}\cdot C^A_\mathbf{r}(\mathbf{p}-\mathbf{r},d),\\
\hspace{-2mm} {\mathbf{w}_2(\mathbf{p},\mathbf{r})}\cdot C^A_\mathbf{r}(\mathbf{p}-\mathbf{r},d-1),\\
\hspace{-2mm} {\mathbf{w}_3(\mathbf{p},\mathbf{r})}\cdot C^A_\mathbf{r}(\mathbf{p}-\mathbf{r},d+1),\\
\hspace{-2mm} {\mathbf{w}_4(\mathbf{p},\mathbf{r})}\cdot \max\limits_{i}{C^A_\mathbf{r}(\mathbf{p}-\mathbf{r},i)}.
\end{array}\right.
\end{array}
\label{EQ:sgmlayer}
\end{equation}
This is different from the SGM in three ways. First, we make the user-defined parameters learnable and add them as penalty coefficients/weights of the matching cost terms. These weights would therefore be adaptive and more flexible at different locations for different situations. Second, we replace the first/external minimum selection in Eq. (\ref{EQ:sgm_agg}) with a weighted sum, without any loss in accuracy. 
This change was proven effective in \cite{springenberg2014striving}, where convolutions with strides were used to replace the max-pooling layers to get an all convolutional network without loss of accuracy. Third, the internal/second minimum selection is changed to a maximum. This is because the learning target in our models is to maximize the probabilities at the ground truth depths instead of minimizing the matching costs. Since $\max\limits_{i}{C^A_\mathbf{r}(\mathbf{p}-\mathbf{r},i)}$ in Eq. (\ref{EQ:sgmlayer}) can be shared by $C^A_\mathbf{r}(\mathbf{p},d)$ for $d$ different locations, here, we do not use another weighted summation to replace it in order to reduce the computational complexity.

For both Eq. (\ref{EQ:sgm_agg}) and Eq. (\ref{EQ:sgmlayer}), the values of $C^A_\mathbf{r}(\mathbf{p},d)$  increase along the path, which may lead to very large values. We normalize the weights of the terms to avoid such a problem. 
This leads to our new semi-global aggregation:
\begin{equation}
\setlength{\abovedisplayskip}{5pt}
\setlength{\belowdisplayskip}{4pt}
 \begin{array}{rll}
C^A_\mathbf{r}(\mathbf{p},d)&\hspace{-1.7mm}=&\hspace{-4.2mm}\text{sum}\left\{\begin{array}{l}
\hspace{-2mm} {\mathbf{w}_0(\mathbf{p},\mathbf{r})} \cdot C(\mathbf{p},d)\\
\hspace{-2mm} {\mathbf{w}_1(\mathbf{p},\mathbf{r})}\cdot C^A_\mathbf{r}(\mathbf{p}-\mathbf{r},d),\\
\hspace{-2mm} {\mathbf{w}_2(\mathbf{p},\mathbf{r})}\cdot C^A_\mathbf{r}(\mathbf{p}-\mathbf{r},d-1),\\
\hspace{-2mm} {\mathbf{w}_3(\mathbf{p},\mathbf{r})}\cdot C^A_\mathbf{r}(\mathbf{p}-\mathbf{r},d+1).\\
\hspace{-2mm} {\mathbf{w}_4(\mathbf{p},\mathbf{r})}\cdot \max\limits_{i}{C^A_\mathbf{r}(\mathbf{p}-\mathbf{r},i)}.\\
\end{array}\right.\\[0.9cm]
&s.t.&\sum\limits_{i=0,1,2,3,4}{\mathbf{w}_i(\mathbf{p},\mathbf{r})}=1
\end{array}
\label{EQ:sgmlayer2}
\end{equation}
$C(\mathbf{p},d)$ is known as the cost volume (with a size of $H \times W \times D_{max}\times F$). Same as the traditional SGM \cite{hirschmuller2008stereo}, the cost volume can be sliced into $D_{max}$ slices at the third dimension for each candidate disparity $d$ and each of these slices repeats the aggregation operation of Eq. (\ref{EQ:sgmlayer2}) with the shared weight matrices ($\mathbf{w}_{0...4}$). All the weights $\mathbf{w}_{0...4}$ can be achieved by a guidance subnet (as shown in Fig. \ref{fig:architecture}). Different to the original SGM which aggregates in sixteen directions, in order to improve the efficiency, the proposed aggregations are done in totally four directions (left, right, up and down) along each row or column over the whole image, namely $\mathbf{r}\in\{(0,1), (0,-1), (1, 0), (-1, 0)\}$. 

The final aggregated output $C^A(\mathbf{p})$ is obtained by selecting the maximum between the four directions:
\begin{equation}
\setlength{\abovedisplayskip}{5pt}
\setlength{\belowdisplayskip}{4pt}
C^A(\mathbf{p},d)=\max\limits_{\mathbf{r}}{~C^A_\mathbf{r}(\mathbf{p},d)}
\label{EQ:maxselect}
\end{equation}
The last maximum selection keeps the best message from only one direction. This guarantees that the aggregation effects are not blurred by the other directions.
The backpropagation for $\mathbf{w}$ and $C(\mathbf{p},d)$ in the SGA layer can be done inversely as Eq. (\ref{EQ:sgmlayer2}) (details are available in the Appendix \ref{apdx:backpropagation}.).
Our SGA layer can be repeated several times in the neural network model to obtain better cost aggregation effects (as illustrated in Fig. \ref{fig:architecture}).
\vspace{-2mm}
\subsubsection{Local Aggregation}
We now introduce the local guided aggregation (LGA) layer which aims to refine the thin structures and object edges. Down-sampling and up-sampling are widely used in stereo matching models which blurs thin structures and object edges. The LGA layer learns several guided filters to refine the matching cost and aid in the recovery of thin structure information. The local aggregation follows the cost filter definition \cite{hosni2013fast} (Eq. (\ref{EQ:costfilter})) and can be written as:
\begin{equation}
\setlength{\abovedisplayskip}{5pt}
\setlength{\belowdisplayskip}{3.5pt}
\begin{array}{c}
\begin{array}{lcc}
\hspace{-2mm}C^A(\mathbf{p},d)&\hspace{-2mm}=&\hspace{-2mm}\text{sum}\left\{\begin{array}{l}
\hspace{-2mm}\sum\nolimits_{\mathbf{q}\in N_\mathbf{p}}{\omega_0(\mathbf{p},\mathbf{q}) \cdot C(\mathbf{q},d)},\\
\hspace{-2mm}\sum\nolimits_{\mathbf{q}\in N_\mathbf{p}}{\omega_1(\mathbf{p},\mathbf{q}) \cdot C(\mathbf{q},d-1)},\\
\hspace{-2mm}\sum\nolimits_{\mathbf{q}\in N_\mathbf{p}}{\omega_2(\mathbf{p},\mathbf{q}) \cdot C(\mathbf{q},d+1)}.
\end{array}\right.
\end{array}\\[6mm]
\hspace{-2mm}s.t.~~\sum\limits_{\mathbf{q}\in N_\mathbf{p}}{\omega_0(\mathbf{p},\mathbf{q})
+\omega_1(\mathbf{p},\mathbf{q})
+\omega_2(\mathbf{p},\mathbf{q})}=1
\end{array}
\label{EQ:localagg}
\end{equation}
 Different slices (totally $D_{max}$ slices) of cost volume share the same filtering/aggregation weights in LGA. This is the same as the original cost filter framework \cite{hosni2013fast} and the SGA (Eq.(\ref{EQ:sgmlayer2})) in this paper. 
While, different with the traditional cost filter \cite{hosni2013fast} which uses a $K\times K$ filter kernel to filter the cost volume in a $K\times K$ local/neighboor region $N_{\mathbf{p}}$, the proposed LGA layer has three $K\times K$ filters ($\omega_0$, $\omega_1$ and $\omega_2$) at each pixel location $\mathbf{p}$ for disparities $d$, $d-1$ and $d+1$ respectively. Namely, it aggregates with a $K\times K \times 3$ weight matrix in a $K\times K$ local region for each pixel location $\mathbf{p}$. The setting of the weight matrix is also similar to \cite{jia2016dynamic}, but, weights and filters are shared during the aggregation as designed in \cite{hosni2013fast}.

\vspace{-1mm}
\subsubsection{Efficient Implementation}
\vspace{-0.5mm}
We use several 2D convolutional layers to build a fast guidance subnet (as illustrated in Fig. \ref{fig:architecture}). The implementation is similar to \cite{zhang2017supplementary}. It uses the reference image as input and outputs the aggregation weights $\mathbf{w}$ (Eq. (\ref{EQ:sgmlayer2})). For a 4D cost volume $C$ with size of $H\times W\times D\times F$ ($H$: height, $W$: width, $D$: max disparity, $F$: feature size), the output of the guidance subnet is split, reshaped and normalized as four $H\times W\times K \times F$ ($K=5$) weight matrices for four directions' aggregation using Eq. (\ref{EQ:sgmlayer2}). Note that aggregations for different disparities corresponding to a slice $d$ share the same aggregation weights. Similarly, the LGA layer need to learn a $H\times W\times 3K^2 \times F$ ($K=5$) weight matrix and aggregates using Eq. (\ref{EQ:localagg}). 

Even though the SGA layer involves an iterative aggregation across the width or the height, the forward and backward can be computed in parallel due to the independence between elements in different feature channels or rows/columns. For example, when aggregating in the left direction, the elements in different channels or rows are independent and can be computed simultaneously. The elements of the LGA layer can also be computed in parallel by simply decomposing it into element-wise matrix multiplications and summations. In order to increase the receptive field of the LGA layer,  we repeat the computation of EQ. (\ref{EQ:localagg}) twice with the same weight matrix, which is similar to \cite{cheng2018depth}.
\vspace{-0.5mm}
\subsection{Network Architecture}
As illustrated in Fig.\ref{fig:architecture}, the GA-Net consists of four parts: the feature extraction block, the cost aggregation for the 4D cost volume, the guidance subnet to produce the cost aggregation weights and the disparity regression. For the feature extraction, we use a stacked hourglass network which is densely connected by concatenations between different layers. The feature extraction block is shared by both left and right views. The extracted features for left and right images are then used to form a 4D cost volume. Several SGA layers are used for the cost aggregation and LGA layers can be implemented before and after the softmax layer of the disparity regression. It refines the thin-structures and compensate for the accuracy loss caused by the down-sampling done for the cost volume. The weight matrices (in Eq.(\ref{EQ:sgmlayer2}) and Eq.(\ref{EQ:localagg})) are generated by an extra guidance subnet which uses the reference view (\eg the left image) as input. The guidance subnet consists of several fast 2D convolutional layers and the outputs are reshaped and normalized into required weight matrices for these GA layers.\footnote{The parameter settings of ``GA-Net-15'' used in our experiments are detailed in the Appendix \ref{apdx:arch}.}

\subsection{Loss Function}
We adopt the smooth
$L_1$ loss function to train our models. Smooth $L_1$ is robust at disparity discontinuities and has low sensitivity to outliers or noises,  as compared to $L_2$ loss. The loss function for training our models is defined as:
\begin{equation}
\setlength{\abovedisplayskip}{3pt}
\setlength{\belowdisplayskip}{3pt}
\begin{array}{rll}
\hspace{-1.5mm}L(\hat d, d)&\hspace{-1.5mm}=&\hspace{-2.5mm}  \frac{1}{N}\sum\limits_{n=1}^{N}{l(|\hat d- d|)}\\[0.4cm]
\hspace{-1.5mm}l(x)&\hspace{-1.5mm}=&\hspace{-3mm}\left\{\begin{array}{ll}
\hspace{-1.5mm} x-0.5, & x \ge 1\\
\hspace{-1.5mm} x^2/2, & x  < 1
\end{array}\right.
\end{array}
\label{EQ:loss}
\end{equation}
where, $|\hat d -d|$ measures the absolute error of the disparity predictions, $N$ is the number of valid pixels with ground truths for training.

For the disparity estimation, we employ the disparity regression proposed in \cite{kendall2017end}:
\begin{equation}
\setlength{\abovedisplayskip}{3pt}
\setlength{\belowdisplayskip}{3pt}
\hat d = \sum\limits_{d=0}^{D_{max}}{d\times\sigma(-C^A(d))}
\label{EQ:regression}
\end{equation}

The disparity prediction $\hat d$ is the sum of each disparity candidate weighted by its probability. The probability of each disparity $d$ is calculated  after cost aggregation via the softmax operation $\sigma(\cdot)$. The disparity regression is shown more robust than classification based methods and can generate sub-pixel accuracy.

\begin{table*}[t]
	\setlength{\abovecaptionskip}{5pt}
	\setlength{\belowcaptionskip}{-5pt}
	\vspace{-0.5mm}
	\centering
	\footnotesize
	\caption{\small Evaluations of GA-Nets with different settings. Average end point error (EPE) and threshold error rate are used for evaluations.}
	\begin{tabular}{c |C{2.5cm} C{1.75cm} C{1.75cm} ||cc | c}
		\hline
		Feature Extraction & \multicolumn{3}{|c|}{Cost Aggregation} & \multicolumn{2}{c|}{Scene Flow} & KITTI 2015 \\
		{Stacked Block}& Densely Concatenate &SGA Layer& LGA Layer& EPE Error & Error Rates (\%) & ~~~~Error Rates (\%)~~~~\\
		\hline\hline
		& & & &1.26 & 13.4&3.39\\
		$\surd$&       &    &  &1.19& 13.0 &3.31\\
		$\surd$&$\surd$&    &  &1.14  &12.5 &3.25\\
		$\surd$&$\surd$& +1 &  &1.05  &11.7 &3.09\\
		$\surd$&$\surd$& +2 &  &0.97  &11.0 &2.96\\
		$\surd$&$\surd$& +3 &  &0.90  & 10.5 &2.85\\
		$\surd$&$\surd$& +4 &  &0.89  & 10.4 &2.83\\
		$\surd$&$\surd$& +3 & $\surd$ & {\bf 0.84} &{\bf 9.9} &{\bf 2.71}\\
		\hline
	\end{tabular}
	\label{tab:ablation}%
	\vspace*{-0.075in}
\end{table*}
\section{Experiments}
In this section, we evaluate our GA-Nets with different settings using  Scene Flow \cite{mayer2016large} and KITTI \cite{kitti2012,kitti2015} datasets.
We implement our architectures
using pytorch or caffe \cite{jia2014caffe} (only for real-time models' implementation). All models are optimized with  Adam
($\beta_1 = 0.9$, $\beta_2 = 0.999$). We train with a batch size of 16 on eight GPUs using  $240\times576$
random crops from the input images. The maximum of the disparity is set as 192. Before training,
we normalize each channel of the image by subtracting their means and dividing their standard deviations. We train the model on Scene Flow dataset for 10 epochs with a constant learning rate
of 0.001. For the KITTI datasets, we fine-tune the models
pre-trained on Scene Flow dataset for a further 640 epochs.  The learning rate for fine-tuning begins at 0.001
for the first 300 epochs and decreases to 0.0001 for the remaining epochs. 


\subsection{Ablation Study}

We evaluate the performance of GA-Nets with different settings, including different architectures and different number (0-4) of GA layers. As listed in Table \ref{tab:ablation}, The guided aggregation models significantly outperform the baseline setting which only has 3D convolutional layers for cost aggregation.  The new architectures for feature extraction and cost aggregation improve the accuracy by 0.14\% on KITTI dataset and 0.9\% on Scene Flow dataset. Finally, the best setting of GA-Net with three SGA layers and one LGA layer gets the best 3-pixel threshold error rate of 2.71\% on KITTI 2015 validation set. It also achieves the best average EPE of 0.84 pixel and the best 1-pixel threshold error rate of 9.9\% on the Scene Flow test set.

\subsection{Effects of Guided Aggregations}
In this section, we compare the guided aggregation strategies with other matching cost aggregation methods. We also analyze the effects of the GA layers by observing the post-softmax probabilities output by different models.

 Firstly, our proposed GA-Nets are compared with the cost aggregation architectures in GC-Net (with nineteen 3D convolutions) and PSMNet (with twenty-five 3D convolutions).  We fixed the feature extraction architecture as proposed above. As shown in Table \ref{tab:aggregation}, GA-Nets have fewer parameters, run at a faster speed and achieve better accuracy. \Eg, with only two GA layers and two 3D convolutions, our GA-Net-2 outperforms the GC-Net by 0.29 pixel in average EPE. Also, the GA-Net-7 with three GA layers and seven 3D convolutions outperforms the current best PSMNet \cite{chang2018pyramid} which has twenty-five 3D convolutional layers.

 \begin{figure}[t]
\setlength{\abovecaptionskip}{10pt}
\setlength{\belowcaptionskip}{-10pt}
\vspace{-4mm}
\centering
	\begin{overpic}[width=0.75\linewidth,height=0.6\linewidth]{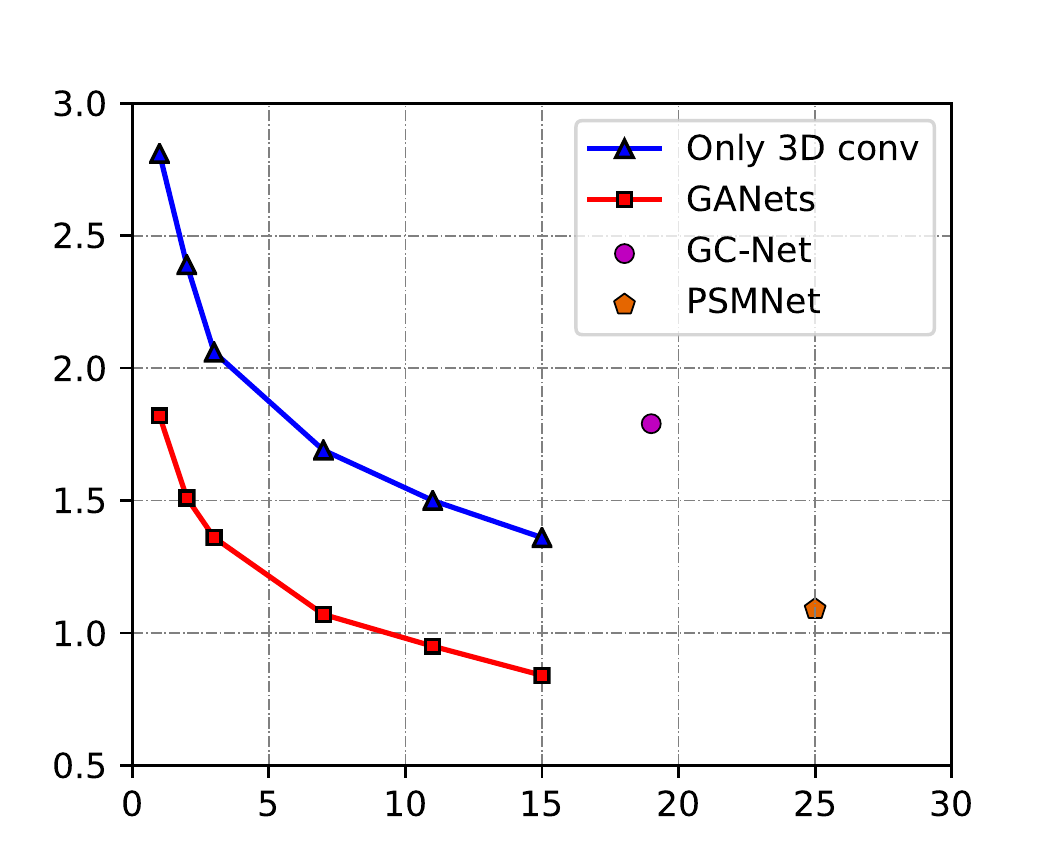}
        \put(-2,23){\footnotesize \color{black}{\begin{turn}{90}End point error\end{turn}}}
        \put(30,-2){\footnotesize \color{black}{Number of 3D conv}}
        \end{overpic}
\caption{\small Illustration of the effects of guided aggregations. GA-Nets are compared with the same architectures without GA Layers. Evaluations are on Scene Flow dataset using average EPE.}
\label{fig:cmp}
\end{figure}

 We also study the effects of the GA layers by comparing with the same architectures without GA steps.
  These baseline models ``GA-Nets$^*$'' have the same network architectures and all other settings except that there is no GA layer implemented. As shown in Fig. \ref{fig:cmp}, for all these models, GA layers have significantly improved the models' accuracy (by 0.5-1.0 pixels in average EPE).
  For example, the GA-Net-2 with two 3D convolutions and two GA layers produces lower EPE (1.51) compared with GA-Net$^*$-11 (1.54) which utilizes eleven 3D convolutions. This implies that two GA layers are more effective than nine 3D convolutional layers.

\begin{table}[t]
	\setlength{\abovecaptionskip}{5pt}
	\setlength{\belowcaptionskip}{-5pt}
	\centering
	\footnotesize
	\caption{\small Comparisons of different cost aggregation methods. Average end point error (EPE) and 1-pixel threshold error rate are used for evaluations on Scene Flow dataset.}
	\begin{tabular}{C{1.4cm}  C{1.1cm} C{0.7cm} C{0.75cm} C{0.9cm} C{0.9cm}}
		\hline
		 \multirow{2}[0]{*}{\tabincell{c}{Models}} & \multirow{2}[0]{*}{\tabincell{c}{3D Conv\\Number}}&  \multirow{2}[0]{*}{\tabincell{c}{Param}}&  \multirow{2}[0]{*}{\tabincell{c}{Time(s)}} & \multirow{2}[0]{*}{\tabincell{c}{EPE\\Error}}& \multirow{2}[0]{*}{\tabincell{c}{Error\\Rates}}\\
        &  & &  & & \\
		\hline\hline
        GC-Net& 19 & 2.9M&4.4 &1.80 &15.6\\
        PSMNet & 25 &3.5M & 2.1& 1.09 &12.1\\
        \hline
        GA-Net-1 & 1 &0.5M & 0.17& 1.82 &16.5\\
        GA-Net-2 & 2 &0.7M & 0.35& 1.51& 15.0\\
        GA-Net-3 & 3 &0.8M & 0.42& 1.36& 13.9\\
        GA-Net-7 & 7 &1.3M & 0.62 & 1.07&11.9\\
        GA-Net-11 & 11 & 1.8M&0.95  &0.95&10.8\\
        GA-Net-15 & 15 & 2.3M&1.5& {\bf 0.84}&{\bf 9.9}\\	
		\hline
	\end{tabular}
	\label{tab:aggregation}%
	\vspace*{-0.095in}
\end{table}

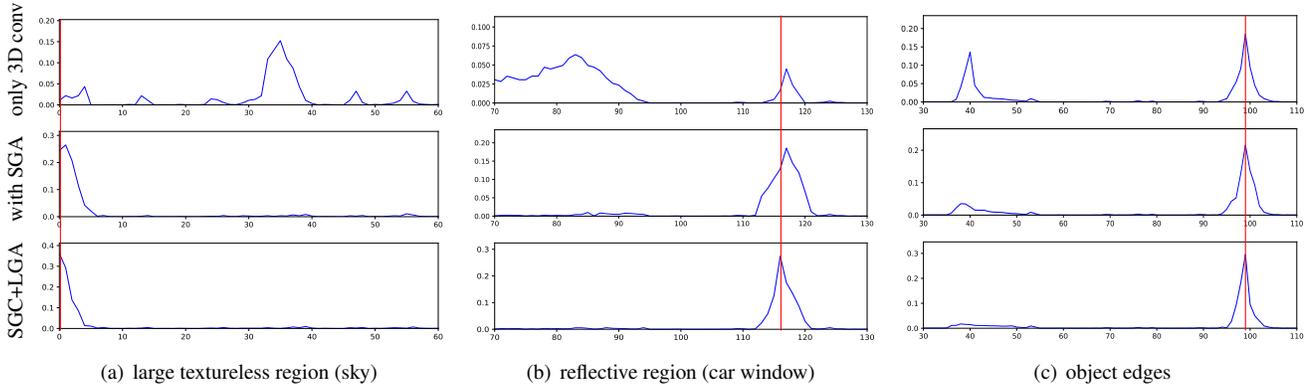
\begin{figure*}[t]
 \setlength{\abovecaptionskip}{0pt}
\setlength{\belowcaptionskip}{-4pt}
 \vspace{-2mm}
 \centering
 \hspace{1mm}
 \subfigure[large textureless region (sky)]{
 	\begin{overpic}[width=0.32\linewidth,height=0.26\linewidth]{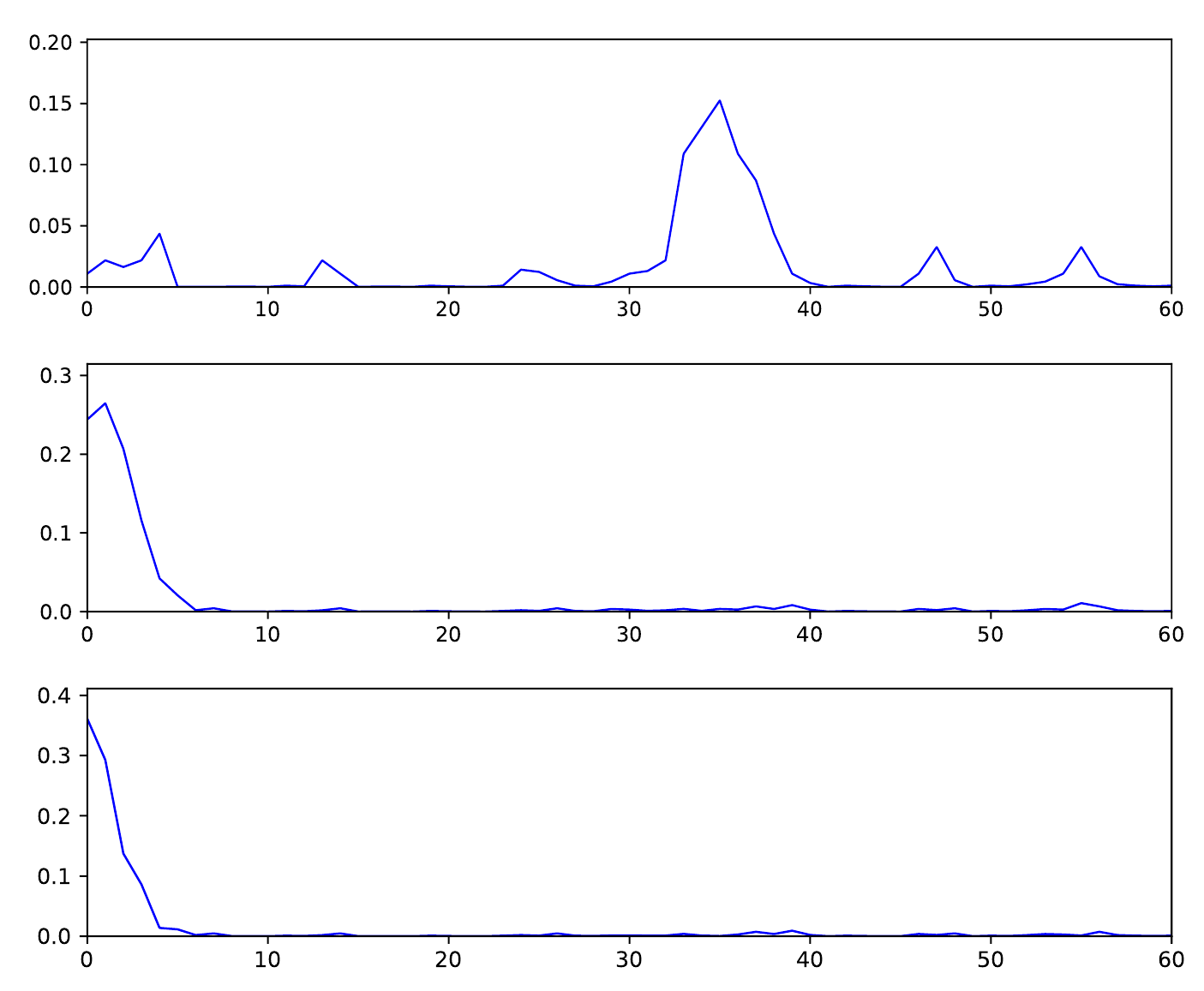}
         \put(-5,55){\footnotesize \color{black}{\begin{turn}{90}only 3D conv\end{turn}}}
          \put(-5,30){\footnotesize \color{black}{\begin{turn}{90} with SGA \end{turn}}}
           \put(-5,2){\footnotesize \color{black}{\begin{turn}{90}SGC+LGA \end{turn}}}
        \put(7.5,4){\color{red}\line(0,1){74}}
     \end{overpic}
 \label{subfig:sky}
 }
  \hspace{-2mm}
  \subfigure[reflective region (car window)]{
  \begin{overpic}[width=0.32\linewidth,height=0.26\linewidth]{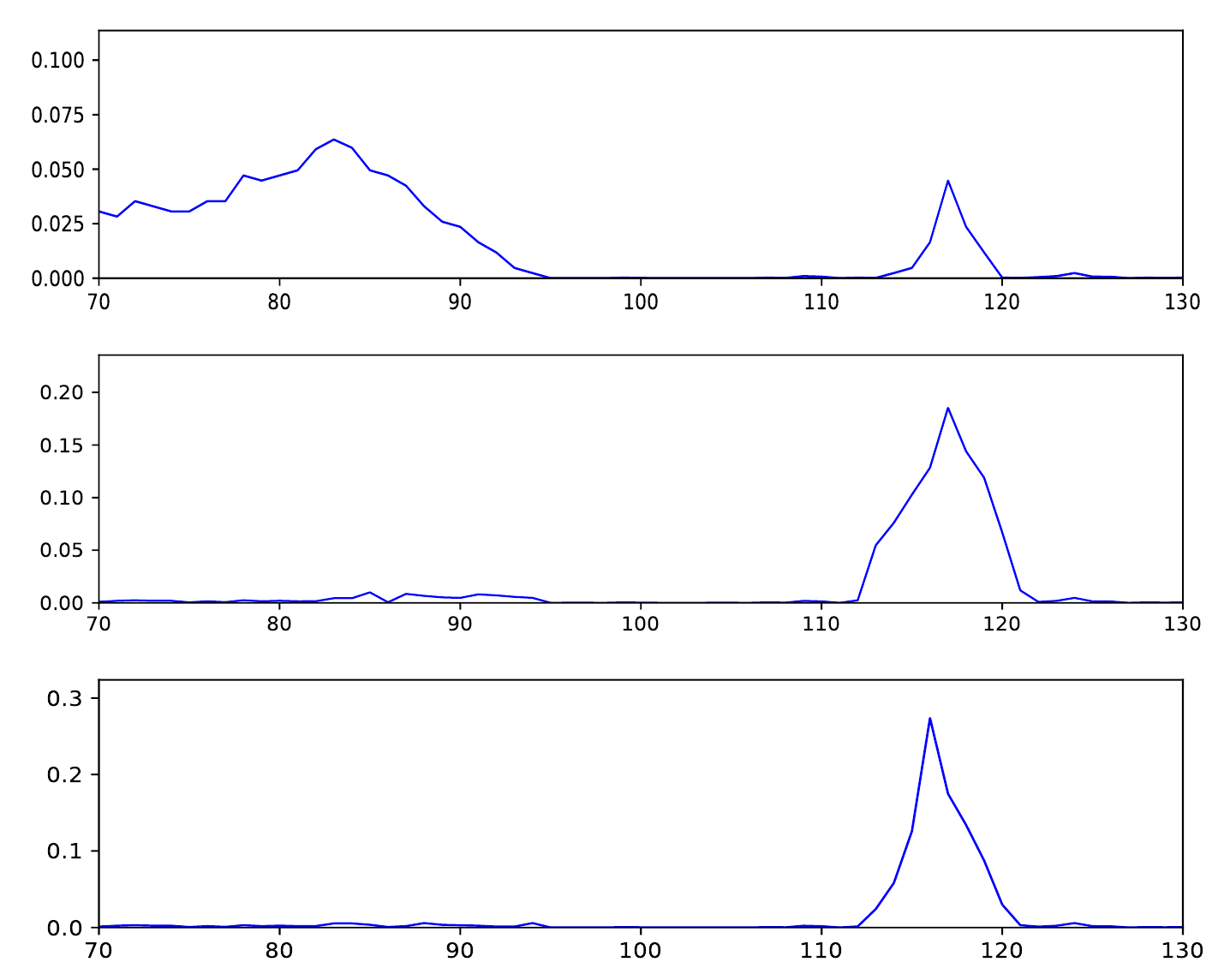}
        \put(76.2,4){\color{red}\line(0,1){75}}
     \end{overpic}
 \label{subfig:window}
 }
  \hspace{-2mm}
  \subfigure[object edges]{
  \begin{overpic}[width=0.32\linewidth,height=0.26\linewidth]{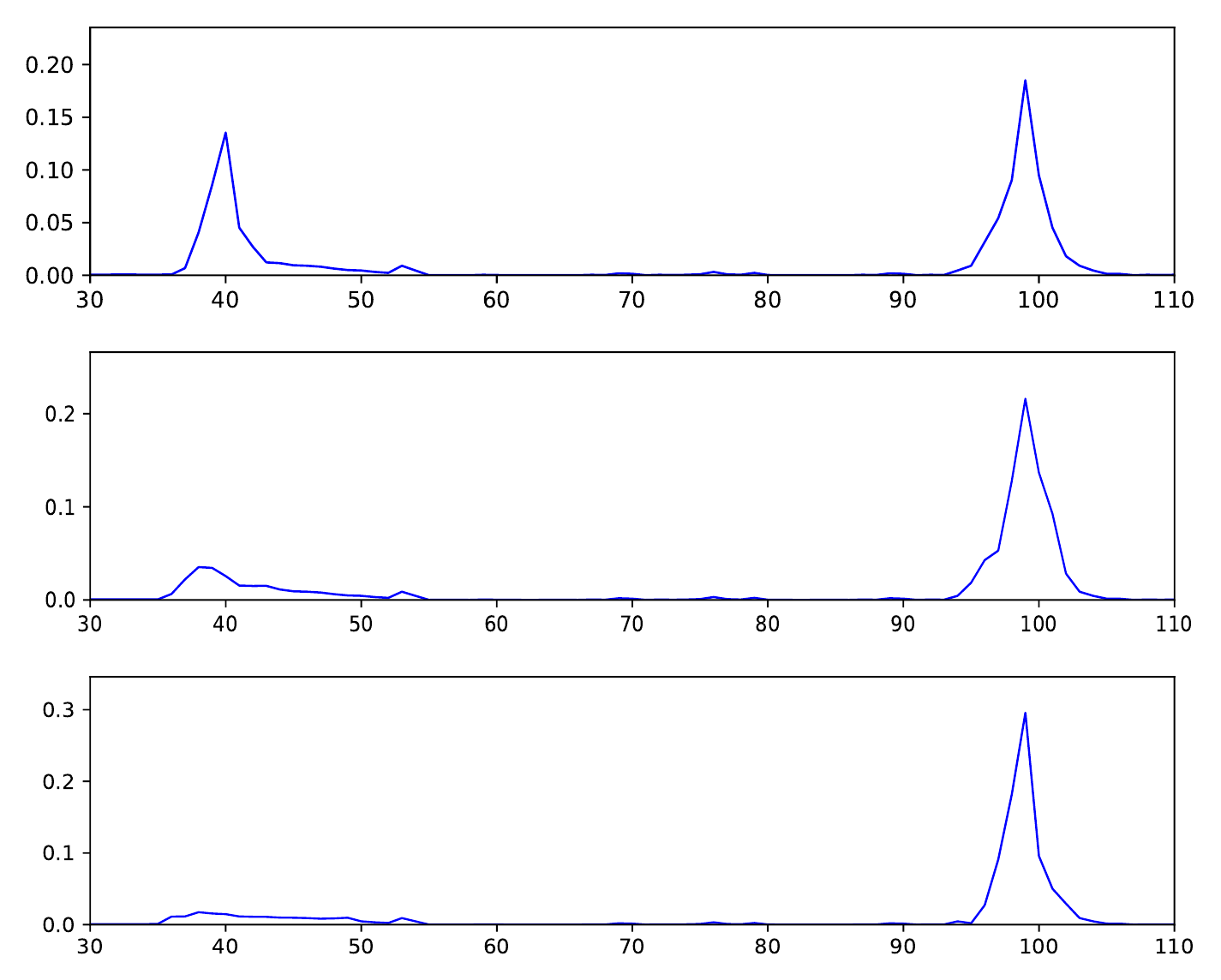}
        \put(84,4){\color{red}\line(0,1){75}}
     \end{overpic}
 \label{subfig:edges}
 }
 \caption{\small Post-softmax probability distributions with respect to disparity values. Red lines illustrate the ground truth disparities. Samples are selected from three challenging regions: (a) the large smooth region (sky), (b) the reflective region from one car window and (c) one region around the object edges. The {\em first row} shows the probability distributions without GA layers. The {\em second row} shows the effects of semi-global aggregation (SGA) layers and the {\em last row} is the refined probabilities with one extra local guided aggregation (LGA) layer.}
 \label{fig:effects}
 \end{figure*}

Finally, in order to observe and analyze the effects of GA layers, in Fig. \ref{fig:effects}, we plot the post-softmax probabilities with respect to a range of candidate disparities. These probabilities are directly used for disparity estimation using Eq. (\ref{EQ:regression}) and can reflect the effectiveness of the cost aggregation strategies. The data samples are all selected from some challenging regions, such as a large textureless region (sky), the reflective region (window of a car) and pixels around the object edges. Three different models are compared.  The first model (first row of Fig. \ref{fig:effects}) only has 3D convolutions (without any GA layers), the second model (second row of Fig. \ref{fig:effects}) has SGA layers and the last model (last row of Fig. \ref{fig:effects}) has both SGA layers and LGA layer.

As illustrated in Fig. \ref{subfig:sky}, for large textureless regions, there would be a lot of noise since there is no any distinctive features in these regions for correct matching. The SGA layers successfully suppress these noise in the probabilities by aggregating surrounding matching information. The LGA layer further concentrates the probability peak on the ground truth value. It could refine the matching results.
Similarly, in the sample of reflective region (Fig. \ref{subfig:window}), the SGA and LGA layers correct the wrong matches and concentrate the peak on the correct disparity value. For the samples around the objects edges (Fig. \ref{subfig:edges}), there are usually two peaks in the probability distribution which are influenced by the background and the foreground respectively. The SGA and LGA use spatial aggregation along with appropriate maximum selection to cut down the aggregation of the wrong matching information from the background and therefore suppress the false probability peak appeared at the background's disparity value.

 \subsection{Comparisons with SGMs and 3D Convolutions}
   The SGA layer is a differentiable approximation of the SGM \cite{hirschmuller2008stereo}. 
 But, it produces far better results compared with both the original SGM with handcrafted features and the MC-CNN \cite{zbontar2015computing} with CNN based features (as shown in Table \ref{tab:kitti2015}). 
 This is because 1) SGA does not have any user-defined parameters that are all learned in an end-to-end fashion.
 2) The aggregation of SGA is fully guided and controlled by the weight matrices. The guidance subnet learns effective geometrical and contextual knowledge to control the directions, scopes and strengths of the cost aggregations.
 
\begin{figure}[t]
 	\setlength{\abovecaptionskip}{-2pt}
 	\setlength{\belowcaptionskip}{-10pt}
 	\vspace{-2mm}
 	\centering
 	\subfigure[Input view]{
 		\includegraphics[width=0.48\linewidth,height=0.15\linewidth]{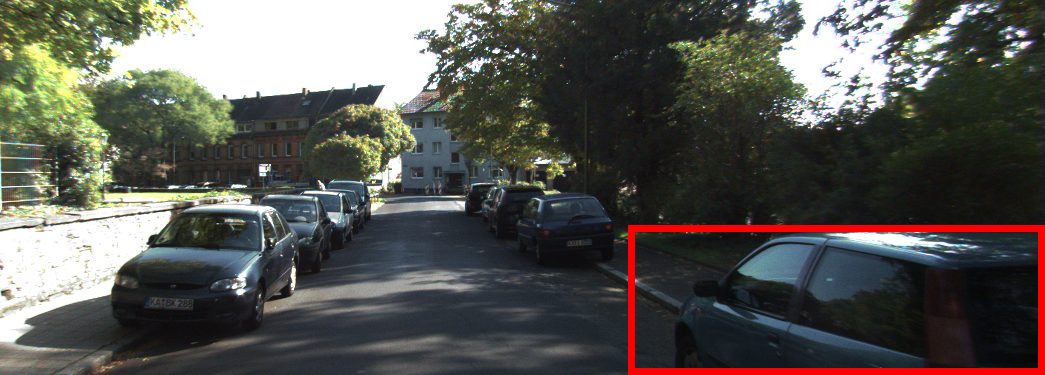}
 	}
 	\hspace{-2.5mm}
 	\subfigure[large textureless region]{
 		\includegraphics[width=0.48\linewidth,height=0.15\linewidth]{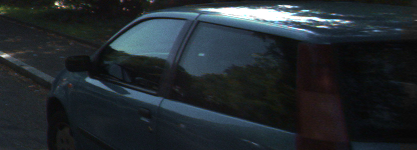}
 	}\\[-2.75mm]
 	\subfigure[result of traditional SGM]{
 		\includegraphics[width=0.48\linewidth,height=0.15\linewidth]{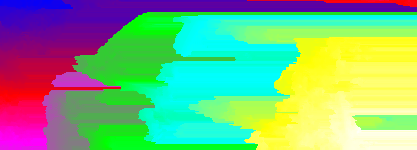}
 	}
 	\hspace{-2.5mm}
 	\subfigure[result of our GA-Net-15]{
 		\includegraphics[width=0.48\linewidth,height=0.15\linewidth]{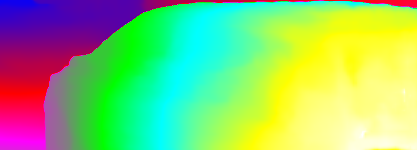}
 	}

 	\caption{\small Comparisons with traditional SGM. More results and comparisons are avaiable at \href{http://www.cvlibs.net/datasets/kitti/eval_scene_flow_detail.php?benchmark=stereo&result=59cfbc4149e979b63b961f9daa3aa2bae021eff3}{GA-Net-15} and \href{http://www.cvlibs.net/datasets/kitti/eval_scene_flow_detail.php?benchmark=stereo&result=474922eb673df9273b4f6db7e87ba12b21f604c0}{SGM}.}
 	\label{fig:sgmCMP}
 	\vspace*{-0.075in}
 \end{figure}

 Moreover, compared with original SGM, most  of the fronto-parallel approximations in large textureless  regions have been avoided. (Example is in Fig. \ref{fig:sgmCMP}.) 
 This might be benefited from:  1) the use of the soft weighted sum in Eq. (\ref{EQ:sgmlayer2}) (instead of the hard min/max selection in  Eq. (\ref{EQ:sgm_agg})); and 2) the regression loss of Eq. (\ref{EQ:regression}) which helps achieve the subpixel accuracy.

Our SGA layer is also more efficient and effective than the 3D convolutional layer.
This is because the 3D convolutional layer could only aggregate in a local region restricted by the kernel size. As a result, a series of 3D convolutions along with encoder and decoder architectures are indispensable in order to achieve good results. As a comparison, our SGA layer aggregates semi-globally in a single layer which is more efficient.
Another advantage of the SGA is that the aggregation's direction, scope and strength are fully guided by variable weights according to different geometrical and contextual information in different locations. \Eg, the SGA behaves totally different in the occlusions and the large smoothness regions. But, the 3D convolutional layer has fixed weights and always perform the same for all locations in the whole image.



%

\begin{figure*}[t]
 \setlength{\abovecaptionskip}{-7pt}
\setlength{\belowcaptionskip}{-10pt}
 \vspace{-2mm}
 \centering
 \begin{minipage}[b]{0.33\linewidth}
 \includegraphics[width=1\linewidth]{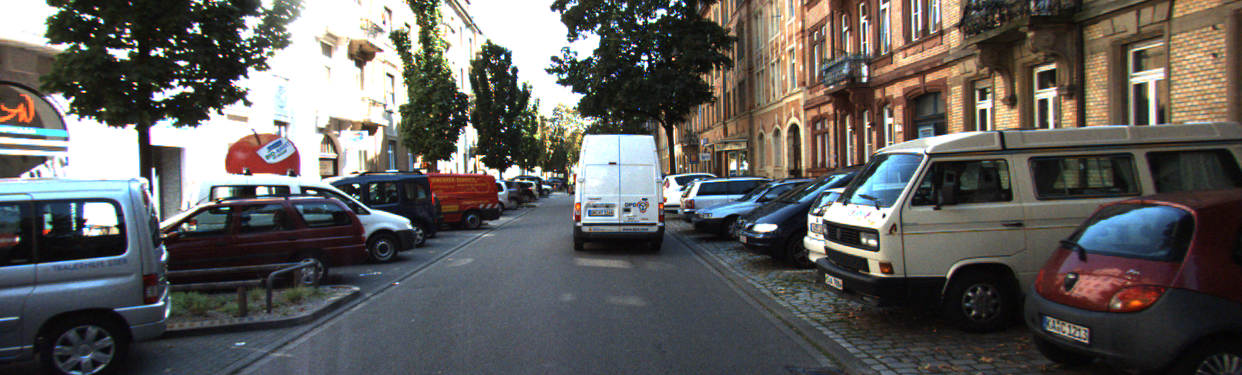}\\
 \includegraphics[width=1\linewidth]{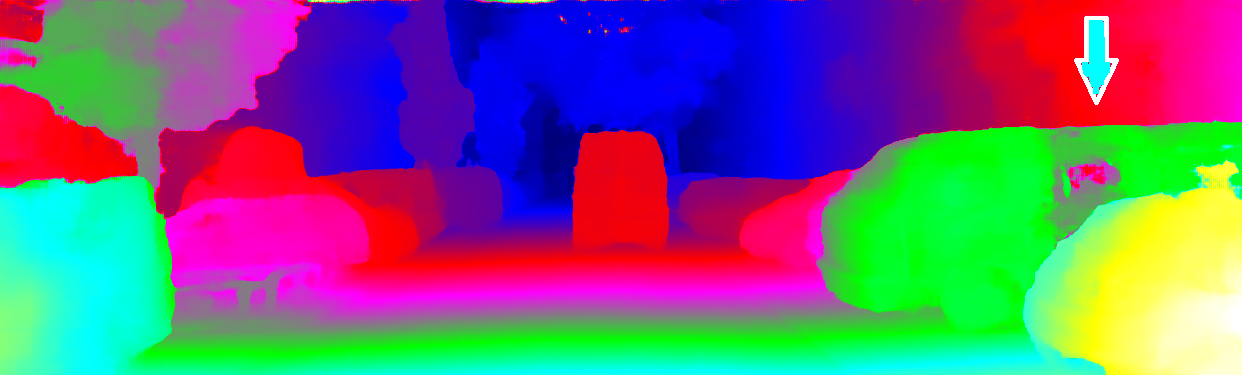}\\
 \includegraphics[width=1\linewidth]{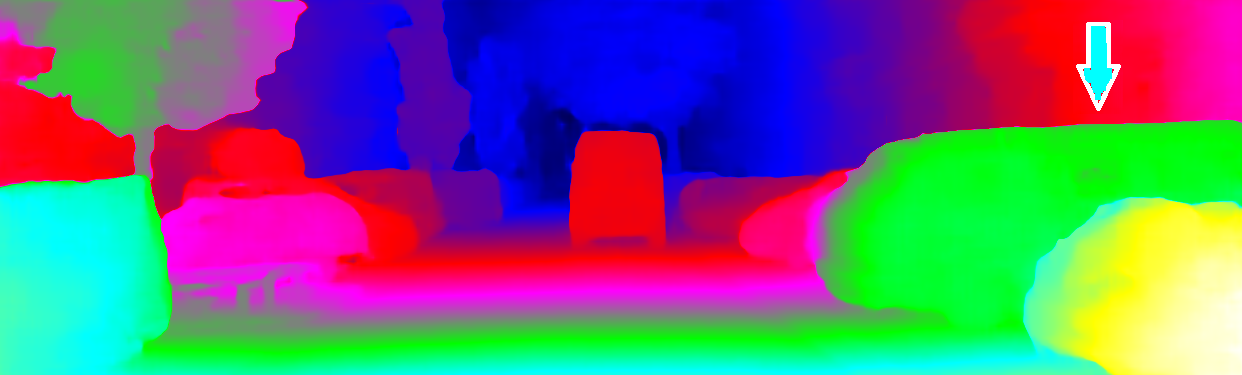}\\
 \includegraphics[width=1\linewidth]{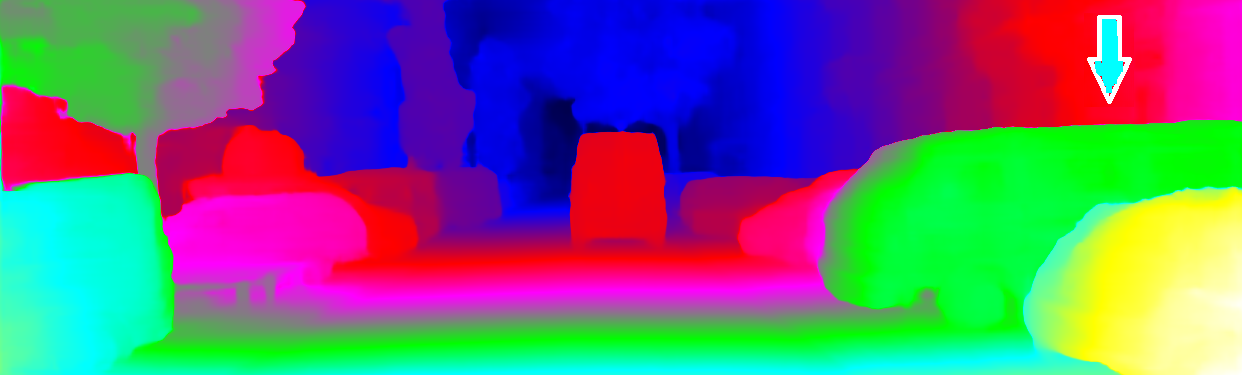}\\
 \end{minipage}
\hspace{-1mm}
 \begin{minipage}[b]{0.33\linewidth}
 \includegraphics[width=1\linewidth]{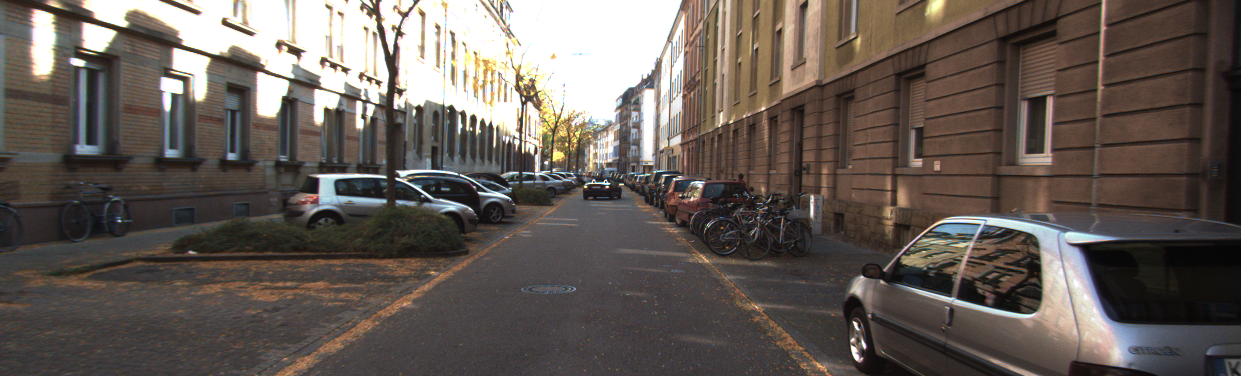}\\
 \includegraphics[width=1\linewidth]{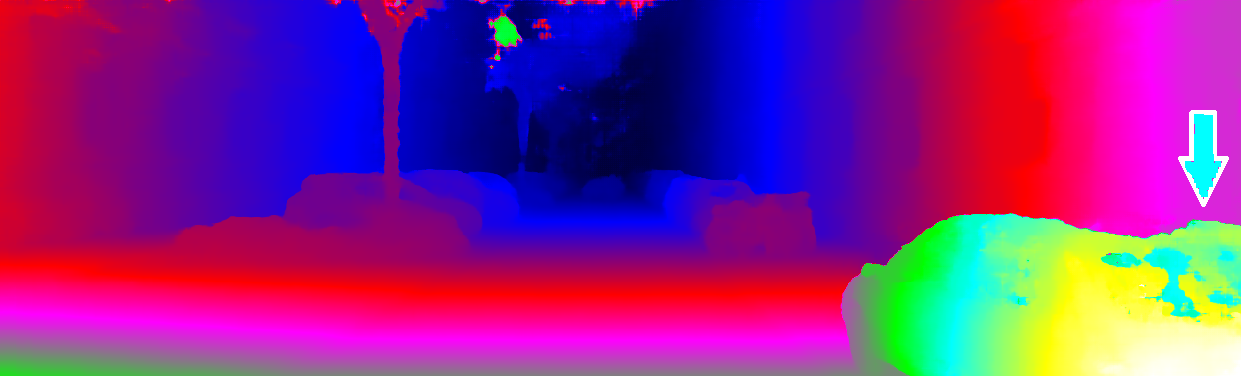}\\
 \includegraphics[width=1\linewidth]{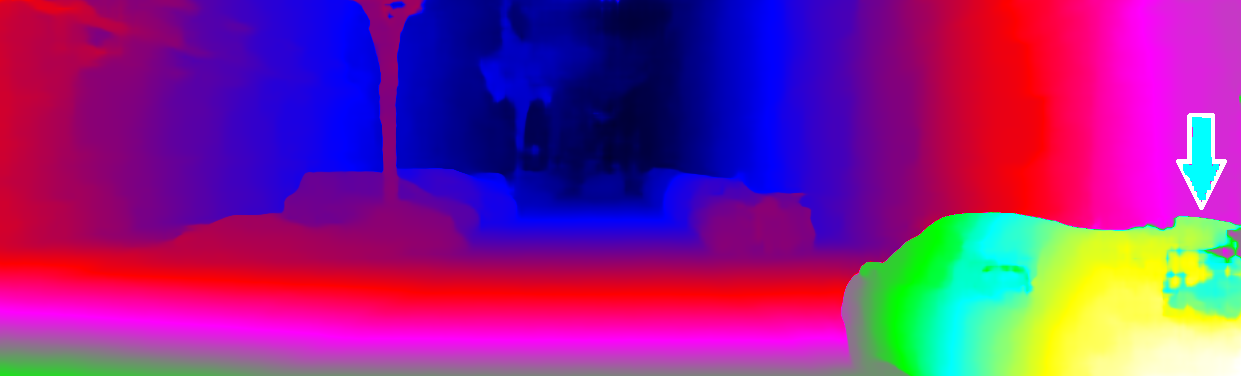}\\
 \includegraphics[width=1\linewidth]{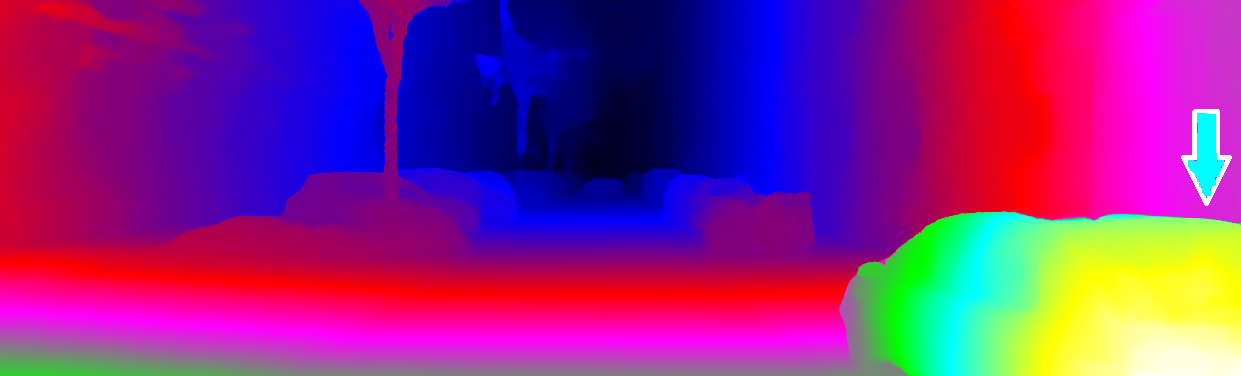}\\
 \end{minipage}
\hspace{-1mm}
 \begin{minipage}[b]{0.33\linewidth}
 \includegraphics[width=1\linewidth]{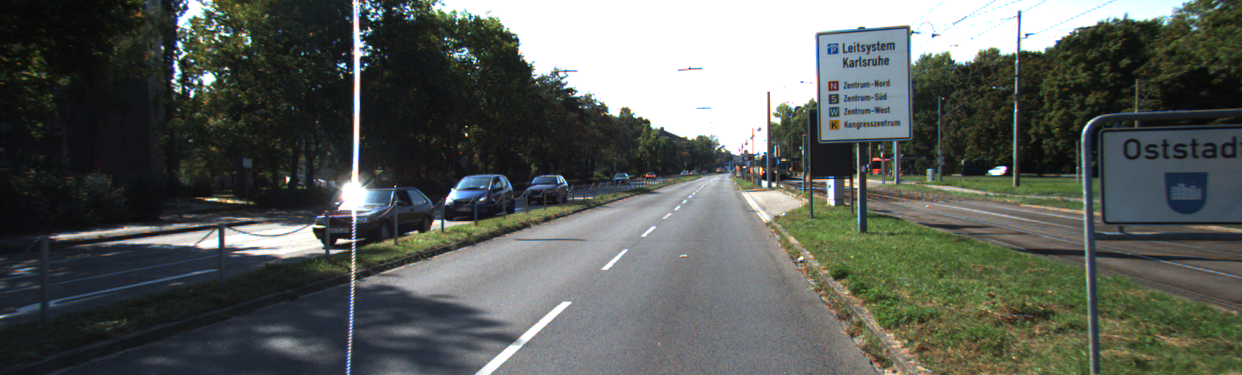}\\
 \includegraphics[width=1\linewidth]{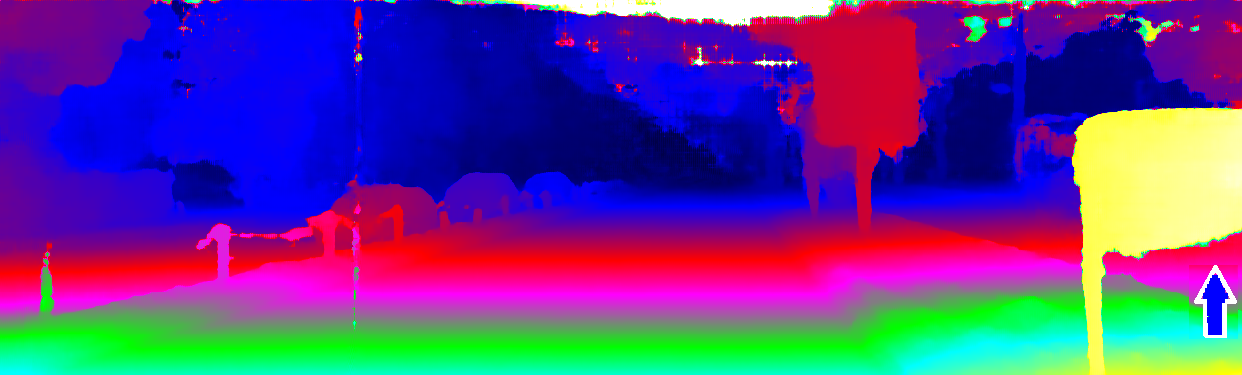}\\
 \includegraphics[width=1\linewidth]{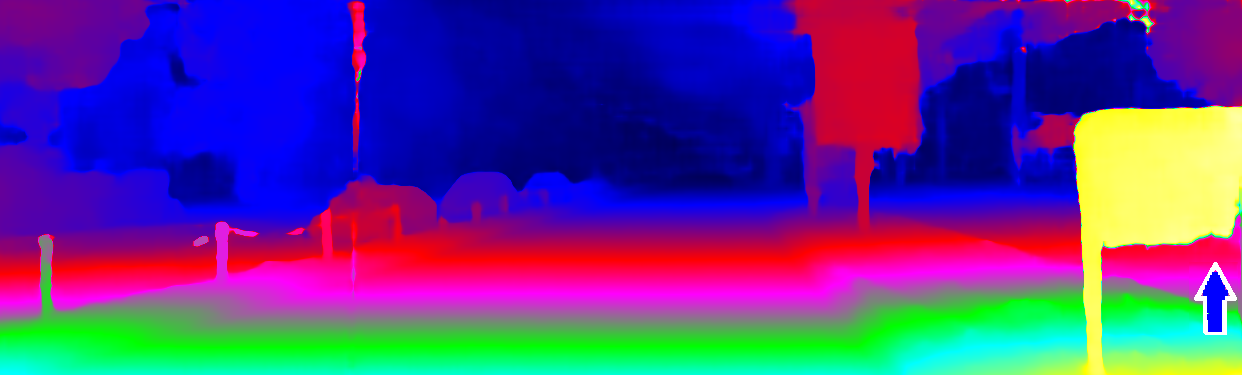}\\
 \includegraphics[width=1\linewidth]{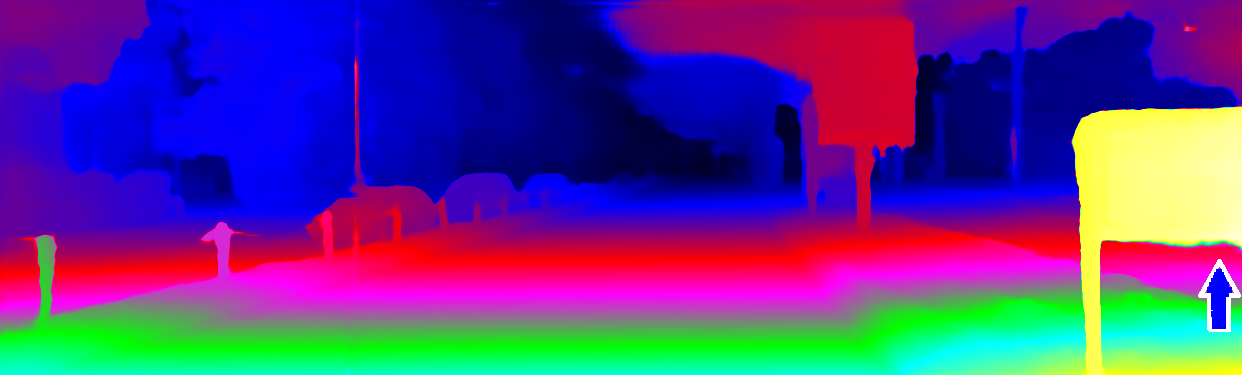}\\
 \end{minipage}
 \caption{\small Results visualization and comparisons. {\em First row:} input image. {\em Second row:} Results of GC-Net\cite{kendall2017end}. {\em Third row:} Results of PSMNet\cite{chang2018pyramid}. {\em Last row:} Results of our GA-Net. Significant improvements are pointed out by blue arrows. The guided aggregations can effectively aggregate the disparity information to the large textureless regions (\eg the cars and the windows) and give precise estimations. It can also aggregate the object knowledge and preserve the depth structure very well (last column). }
 \label{fig:resultsCMP}
 \end{figure*}
\begin{table}[t]
	\setlength{\abovecaptionskip}{5pt}
	\setlength{\belowcaptionskip}{0pt}
	\centering
	\vspace{-1mm}
	\footnotesize
	\caption{\small Comparisons with existing real-time algorithms}
	\begin{tabular}{C{1.5cm}  C{1.72cm} C{1.4cm} C{1.8cm}}
		\hline
		{Methods}  & {End point error} & Error rates & Speed (fps)\\
		\hline\hline
		Our GA-Net & {\bf 0.7 px} & {\bf 3.21 \%} & 15 (GPU)\\
		DispNet\cite{mayer2016large} & 1.0 px & 4.65 \% &22 (GPU)\\
		Toast\cite{ranft2014modeling} & 1.4 px & 7.42 \% &25 (CPU)\\
		\hline
	\end{tabular}
	\label{tab:realtime}%
	\vspace*{-0.095in}
\end{table}
\subsection{Complexity and Real-time Models}
The computational complexity of one 3D convolutional layer is $O(K^3CN)$, where $N$ is the elements number of the output blob. $K$ is the size of the convolutional kernel and $C$ is the channel number of the input blob. As a comparison, the complexity of SGA is $O(4KN)$ or $O(8KN)$ for four or eight-direction aggregations. In GC-Net\cite{kendall2017end} and PSMNet\cite{chang2018pyramid}, $K=3$, $C=32,64$ or $128$ and in our GA-Nets, $K$ is used as 5 (for SGA layer). Therefore, the computational complexity in terms of floating-point operations (FLOPs) of the proposed SGA step is less than 1/100 of one 3D convolutional layer.


The SGA layer are much faster and more effective than 3D convolutions. This allows us to build an accurate real-time model. We implement one caffe\cite{jia2014caffe} version of the GA-Net-1 (with only one 3D convolutional layer and without LGA layers). The model is further simplified by using 4$\times$ down-sampling and up-sampling for cost volume. The real-time model could run at a speed of 15$\sim$20 fps for $300\times1000$ images on a TESLA P40 GPU. We also compare the accuracy of the results with the state-of-the-art real-time models. As shown in Table \ref{tab:realtime}, the real-time GA-Net far outperforms other existing real-time stereo matching models.

 \subsection{Evaluations on Benchmarks}
For the benchmark evaluations, we use the  GA-Net-15 with full settings for evaluations. We compare our GA-Net with the state-of-the-art deep neural network models on the Scene Flow dataset and the KITTI benchmarks.
\subsubsection{Scene Flow Dataset}
The Scene Flow synthetic dataset \cite{mayer2016large} contains
35,454 training and 4,370 testing images. We use the ``final'' set for training and testing. GA-Nets are compared with other state-of-the-art DNN models by evaluating with the average end point errors (EPE) and 1-pixel threshold error rates on the test set. The results are presented in Table \ref{tab:aggregation}. We find that our GA-Net outperforms the state-of-the-arts on both of the two evaluation metrics by a noteworthy margin (2.2\% improvement in error rate and 0.25 pixel improvement in EPE compared with the current best PSMNet\cite{chang2018pyramid}.).
\begin{table}[t]
\setlength{\abovecaptionskip}{5pt}
\setlength{\belowcaptionskip}{0pt}
\vspace{-1mm}
\centering
\footnotesize
\caption{\small Evaluation Results on KITTI 2012 Benchmark}
\begin{tabular}{C{1.55cm}  C{1.1cm} C{1.1cm} C{1.1cm} C{1.1cm}}
\hline
 \multirow{2}[0]{*}{\tabincell{c}{Models}} & \multirow{2}[0]{*}{\tabincell{c}{error rates\\(2 pixels)}}  & \multirow{2}[0]{*}{\tabincell{c}{error rates\\ (3 pixels)}}  & \multirow{2}[0]{*}{\tabincell{c}{Reflective\\regions}} & \multirow{2}[0]{*}{\tabincell{c}{Avg-All\\(end point)}}\\
& & & & \\
\hline\hline
{\bf Our GA-Net} & {\bf 2.18} \% & {\bf 1.36} \%   &{\bf 7.87}\% &{\bf 0.5 px}\\
PSMNet\cite{chang2018pyramid} & 2.44 \% & 1.49 \%  &8.36\% & 0.6 px\\
GC-Net\cite{kendall2017end}  & 2.71 \% & 1.77 \%   &10.80\% & 0.7 px\\
\hspace{-2mm}MC-CNN\cite{zbontar2015computing} & 3.90 \% &2.43 \% &17.09\% &0.9 px\\
\hline
\end{tabular}
\label{tab:kitti2012}%
\vspace*{-0.075in}
\end{table}

\begin{table}[t]
	\setlength{\abovecaptionskip}{5pt}
	\setlength{\belowcaptionskip}{-5pt}
	\centering
	\vspace{1mm}
	\footnotesize
	\caption{\small Evaluation Results on KITTI 2015 Benchmark}
	\begin{tabular}{R{2cm}  C{1.0cm} C{1.0cm} C{1.0cm} C{1.0cm}}
		\hline
		\multirow{2}[0]{*}{\tabincell{c}{Models}} & \multicolumn{2}{c}{Non Occlusion} & \multicolumn{2}{c}{All Areas} \\ 
		& Foreground & Avg All~~ & Foreground & Avg All\\
		\hline\hline
		{\bf Our GA-Net-15}  & {\bf 3.39}\% & {\bf 1.84\%} & {\bf 3.91\%} & {\bf 2.03\%} \\
		PSMNet\cite{chang2018pyramid}  & {4.31}\% & 2.14 \% & {4.62}\% & 2.32\%\\
		GC-Net\cite{kendall2017end}   & {5.58}\% & 2.61\% & {6.16}\% & 2.87\%\\
		SGM-Nets\cite{SGMNet} & 7.43\% & 3.09\% &8.64\% & 3.66\%\\
		MC-CNN\cite{zbontar2015computing}& {7.64}\% &  3.33\% & {8.88}\% & 3.89\% \\
		SGM\cite{hirschmuller2008stereo}   & {11.68\%} & 5.62\% & {13.00}\% & 6.38\%\\
		\hline
	\end{tabular}
	\label{tab:kitti2015}%
	\vspace*{-0.075in}
\end{table}

\subsubsection{KITTI 2012 and 2015 Datasets}

After training on Scene Flow dataset, we use the GA-Net-15 to fine-tune on the KITTI 2015 and KITTI 2012 data sets respectively. The models are then evaluated on the test sets. According to the online leader board, as shown in Table \ref{tab:kitti2012} and Table \ref{tab:kitti2015}, our GA-Net has fewer low-efficient 3D convolutions but achieves better accuracy. It surpasses current best PSMNet 
in all the evaluation metrics. Examples are shown in Fig. \ref{fig:resultsCMP}. The GA-Nets can effectively aggregate the correct matching information into the challenging large textureless or reflective regions to get precise estimations. It also keeps the object structures very well.

\section{Conclusion}

In this paper, we developed much more efficient and effective guided matching cost aggregation (GA) strategies, including the semi-global aggregation (SGA) and the local guided aggregation (LGA) layers for end-to-end stereo matching. The GA layers  significantly improve the accuracy of the disparity estimation in challenging regions, such as occlusions, large textureless/reflective regions and thin structures. The GA layers can be used to replace computationally costly 3D convolutions and get better accuracy.


\appendix
{\centering{\section*{\Large Appendix}}}

\section{\large Backpropagation of SGA}\label{apdx:backpropagation}

The backpropagation for $\mathbf{w}$ and $C(\mathbf{p},d)$ in SGA (Eq.(\ref{EQ:sgmlayer2})) can be computed inversely. Assume the gradient from next layer (max-selection) of Eq. (\ref{EQ:maxselect}) is $\frac{\partial E}{\partial C^A_{r}}$. The backpropagation of SGA can be implemented as: 
\begin{equation}
\frac{\partial E}{\partial C(\mathbf{p},d)}=\sum\limits_{\mathbf{r}}{\frac{\partial E}{\partial C^b_{r}(\mathbf{p},d)}\cdot \mathbf{w}_0(\mathbf{p},\mathbf{r})}.
\end{equation}
\begin{equation}
\begin{array}{l}
\frac{\partial E}{\partial \mathbf{w}_0(\mathbf{p},\mathbf{r})}=\sum\limits_{d}{\frac{\partial E}{\partial C^b_{r}(\mathbf{p},d)}\cdot C(\mathbf{p},d)},\\
\frac{\partial E}{\partial \mathbf{w}_1(\mathbf{p},\mathbf{r})}=\sum\limits_{d}{\frac{\partial E}{\partial C^b_{r}(\mathbf{p},d)}\cdot C^A_\mathbf{r}(\mathbf{p}-\mathbf{r},d)},\\
\frac{\partial E}{\partial \mathbf{w}_2(\mathbf{p},\mathbf{r})}=\sum\limits_{d}{\frac{\partial E}{\partial C^b_{r}(\mathbf{p},d)}\cdot C^A_\mathbf{r}(\mathbf{p}-\mathbf{r},d-1)},\\
\frac{\partial E}{\partial \mathbf{w}_3(\mathbf{p},\mathbf{r})}=\sum\limits_{d}{\frac{\partial E}{\partial C^b_{r}(\mathbf{p},d)}\cdot C^A_\mathbf{r}(\mathbf{p}-\mathbf{r},d+1)},\\
\frac{\partial E}{\partial \mathbf{w}_4(\mathbf{p},\mathbf{r})}=\sum\limits_{d}{\frac{\partial E}{\partial C^b_{r}(\mathbf{p},d)}\cdot \max\limits_{i}{ C^A_\mathbf{r}(\mathbf{p}-\mathbf{r},i)}}.
\end{array}
\end{equation}
where, $\frac{\partial E}{\partial C^b_{\mathbf{r}}}$ is a temporary gradient variable which can be calculated iteratively by (if $d\neq i_{max}$):
\begin{equation}
\begin{array}{rll}
\frac{\partial E}{\partial C^b_{\mathbf{r}}(\mathbf{p},d)} = \frac{\partial E}{\partial C^A_{\mathbf{r}}(\mathbf{p},d)}+ 
\text{sum}\left\{\begin{array}{l}
\hspace{-2mm} { \frac{\partial E}{C^b_\mathbf{r}(\mathbf{p}+\mathbf{r},d)} \cdot \mathbf{w}_1(\mathbf{p}+\mathbf{r},\mathbf{r})},\\
\hspace{-2mm} { \frac{\partial E}{C^b_\mathbf{r}(\mathbf{p}+\mathbf{r},d+1)} \cdot \mathbf{w}_2(\mathbf{p}+\mathbf{r},\mathbf{r})},\\
\hspace{-2mm} { \frac{\partial E}{C^b_\mathbf{r}(\mathbf{p}+\mathbf{r},d-1)} \cdot \mathbf{w}_3(\mathbf{p}+\mathbf{r},\mathbf{r})}.
\end{array}\right.
\end{array}
\label{EQ:backsgm}
\end{equation}
or (if $d= i_{max}$):
\begin{equation}
\begin{array}{rll}
\frac{\partial E}{\partial C^b_{\mathbf{r}}(\mathbf{p},d)} = \frac{\partial E}{\partial C^A_{\mathbf{r}}(\mathbf{p},d)}+ 
\text{sum}\left\{\begin{array}{l}
\hspace{-2mm} { \frac{\partial E}{C^b_\mathbf{r}(\mathbf{p}+\mathbf{r},d)} \cdot \mathbf{w}_1(\mathbf{p}+\mathbf{r},\mathbf{r})},\\
\hspace{-2mm} { \frac{\partial E}{C^b_\mathbf{r}(\mathbf{p}+\mathbf{r},d+1)} \cdot \mathbf{w}_2(\mathbf{p}+\mathbf{r},\mathbf{r})},\\
\hspace{-2mm} { \frac{\partial E}{C^b_\mathbf{r}(\mathbf{p}+\mathbf{r},d-1)} \cdot \mathbf{w}_3(\mathbf{p}+\mathbf{r},\mathbf{r})},\\
\hspace{-2mm} {\sum\limits_i \frac{\partial E}{C^b_\mathbf{r}(\mathbf{p}+\mathbf{r},i)} \cdot \mathbf{w}_4(\mathbf{p}+\mathbf{r},\mathbf{r})}.
\end{array}\right.
\end{array}
\label{EQ:backsgm}
\end{equation}
where $i_{max}$ is the index of $\max\limits_{i}{C^A_\mathbf{r}(\mathbf{p},i)}$ during the forward propagation in Eq. (\ref{EQ:sgmlayer2}).

\section{Details of the Architecture}\label{apdx:arch}
Table \ref{tab:architecture} presents the details of the GA-Net-15 which is used in experiments to produce state-of-the-art accuracy on Scene Flow dataset\cite{mayer2016large} and KITTI benchmarks\cite{kitti2012,kitti2015}.  It has three SGA layers, two LGA layers and fifteen 3D convolutional layers for cost aggregation.

 \begin{table*}[t]
	\renewcommand\arraystretch{0.95}
	\setlength{\abovecaptionskip}{5pt}
	\setlength{\belowcaptionskip}{0pt}
	\centering
	\footnotesize
	\caption{\small Parameters of the network architecture of ``GA-Net-15''}
	\begin{tabular}{C{2cm}|L{6cm} |L{4cm}}
		\hline
		{\bf No.} & {\bf Layer Description} & {\bf Output Tensor}\\
		\hline\hline
		\multicolumn{3}{c}{\bf Feature Extraction} \\
		\hline
		input& normalized image pair as input & H$\times$W$\times$3 \\
		1 & 3$\times$3 conv & H$\times$W$\times$32\\
		2 & 3$\times$3 conv, stride 3  & $\sfrac{1}{3}$H$\times$$\sfrac{1}{3}$W$\times$32\\
		3 & 3$\times$3 conv  & $\sfrac{1}{3}$H$\times$$\sfrac{1}{3}$W$\times$32\\
		4 & 3$\times$3 conv, stride 2  & $\sfrac{1}{6}$H$\times$$\sfrac{1}{6}$W$\times$48\\
		5 & 3$\times$3 conv  & $\sfrac{1}{6}$H$\times$$\sfrac{1}{6}$W$\times$48\\
		6-7 & repeat 4-5 & $\sfrac{1}{12}$H$\times$$\sfrac{1}{12}$W$\times$64\\
		8-9 & repeat 6-7 & $\sfrac{1}{24}$H$\times$$\sfrac{1}{24}$W$\times$96\\
		10-11 & repeat 8-9& $\sfrac{1}{48}$H$\times$$\sfrac{1}{48}$W$\times$128\\
		12 & 3$\times$3 deconv, stride 2 & $\sfrac{1}{24}$H$\times$$\sfrac{1}{24}$W$\times$96\\
		13 & 3$\times$3 conv & $\sfrac{1}{24}$H$\times$$\sfrac{1}{24}$W$\times$96\\
		14-15 & repeat 12-13 & $\sfrac{1}{12}$H$\times$$\sfrac{1}{12}$W$\times$64\\
		16-17 & repeat 12-13 & $\sfrac{1}{6}$H$\times$$\sfrac{1}{6}$W$\times$48\\
		18-19 & repeat 12-13 & $\sfrac{1}{3}$H$\times$$\sfrac{1}{3}$W$\times$32\\
		20-35 & repeat 4-19& $\sfrac{1}{3}$H$\times$$\sfrac{1}{3}$W$\times$32\\
		\hline
	\multirow{2}[0]{*}{\tabincell{c}{concat\\connection}} & 	\multicolumn{2}{l}{\multirow{2}[0]{*}{\tabincell{l}{(9,12), (7,14), (5,16), (3,18), (17,20), (15,22), (13,24), (11,26), (18,25)\\(25,28), (23,30) (21,32), (19,34) }}} \\
	&&\\
		\hline
		{\tabincell{c}{\hspace{-1mm}cost volume}}& by feature concatenation & $\sfrac{1}{3}$H$\times$$\sfrac{1}{3}$W$\times$64$\times$32\\
		\hline
		\multicolumn{3}{c}{\bf Guidance Branch} \\
		\hline
		input& concat 1 and up-sampled 35 as input& H$\times$W$\times$64\\
		(1)&  3$\times$3 conv & H$\times$W$\times$16\\
		(2) & 3$\times$3 conv, stride 3 & $\sfrac{1}{3}$H$\times$$\sfrac{1}{3}$W$\times$32\\
		(3) & 3$\times$3 conv & $\sfrac{1}{3}$H$\times$$\sfrac{1}{3}$W$\times$32\\
		(4) & 3$\times$3 conv (no bn \& relu)& $\sfrac{1}{3}$H$\times$$\sfrac{1}{3}$W$\times$640\\
		(5) &  split, reshape, normalize & $4\times$ $\sfrac{1}{3}$H$\times$$\sfrac{1}{3}$W$\times$5$\times$32\\
		(6) & from (3), 3$\times$3 conv & $\sfrac{1}{3}$H$\times$$\sfrac{1}{3}$W$\times$32\\
		(7) & 3$\times$3 conv (no bn \& relu)& $\sfrac{1}{3}$H$\times$$\sfrac{1}{3}$W$\times$640\\
		(8) &  split, reshape, normalize & $4\times$ $\sfrac{1}{3}$H$\times$$\sfrac{1}{3}$W$\times$5$\times$32\\
		(9)-(11) & from (6), repeat (6)-(8) &$4\times$ $\sfrac{1}{3}$H$\times$$\sfrac{1}{3}$W$\times$5$\times$32\\
		
		(12) &  from (1),  3$\times$3 conv & H$\times$W$\times$16\\
		(13) &  3$\times$3 conv (no bn \& relu) & H$\times$W$\times$75\\
		(14) & split, reshape, normalize & H$\times$W$\times$75\\
		(15)-(17) & from (12), repeat (12)-(14) & H$\times$W$\times$75\\
		\hline
		\multicolumn{3}{c}{\bf Cost Aggregation} \\
		\hline
		input& 4D cost volume & $\sfrac{1}{3}$H$\times$$\sfrac{1}{3}$W$\times$48$\times$64\\
		$[1]$& 3$\times$3$\times$3, 3D conv & $\sfrac{1}{3}$H$\times$$\sfrac{1}{3}$W$\times$48$\times$32\\
		$[2]$&  {\bf SGA} layer: weight matrices from (5) &$\sfrac{1}{3}$H$\times$$\sfrac{1}{3}$W$\times$48$\times$32\\
		$[3]$& 3$\times$3$\times$3, 3D conv & $\sfrac{1}{3}$H$\times$$\sfrac{1}{3}$W$\times$48$\times$32\\
		\multirow{2}[0]{*}{\tabincell{c}{output}} &3$\times$3$\times$3, 3D to 2D conv, upsamping & H$\times$W$\times$193\\
		& softmax, regression, loss weight: 0.2& H$\times$W$\times$1\\
		$[4]$& 3$\times$3$\times$3, 3D conv, stride 2 & $\sfrac{1}{6}$H$\times$$\sfrac{1}{6}$W$\times$48$\times$48\\
		$[5]$& 3$\times$3$\times$3, 3D conv, stride 2 & $\sfrac{1}{12}$H$\times$$\sfrac{1}{12}$W$\times$48$\times$64\\
		$[6]$& 3$\times$3$\times$3, 3D deconv, stride 2 & $\sfrac{1}{6}$H$\times$$\sfrac{1}{6}$W$\times$48$\times$48\\
		$[7]$& 3$\times$3$\times$3, 3D conv & $\sfrac{1}{6}$H$\times$$\sfrac{1}{6}$W$\times$48$\times$48\\
		$[8]$& 3$\times$3$\times$3, 3D deconv, stride 2 & $\sfrac{1}{3}$H$\times$$\sfrac{1}{3}$W$\times$48$\times$32\\
		$[9]$& 3$\times$3$\times$3, 3D conv & $\sfrac{1}{3}$H$\times$$\sfrac{1}{3}$W$\times$48$\times$32\\
		$[10]$& {\bf SGA} layer: weight matrices from (8) &$\sfrac{1}{3}$H$\times$$\sfrac{1}{3}$W$\times$48$\times$32\\
		\multirow{2}[0]{*}{\tabincell{c}{output}} &3$\times$3$\times$3, 3D to 2D conv, upsamping & H$\times$W$\times$193\\
		& softmax, regression, loss weight: 0.6 & H$\times$W$\times$1\\
		
		$[11]$& 3$\times$3$\times$3, 3D conv, stride 2 & $\sfrac{1}{6}$H$\times$$\sfrac{1}{6}$W$\times$48$\times$48\\
		$[12]$& 3$\times$3$\times$3, 3D conv & $\sfrac{1}{6}$H$\times$$\sfrac{1}{6}$W$\times$48$\times$48\\
		$[13]$& 3$\times$3$\times$3, 3D conv, stride 2 & $\sfrac{1}{12}$H$\times$$\sfrac{1}{12}$W$\times$48$\times$64\\
		$[14]$& 3$\times$3$\times$3, 3D deconv, stride 2 & $\sfrac{1}{6}$H$\times$$\sfrac{1}{6}$W$\times$48$\times$48\\
		$[15]$& 3$\times$3$\times$3, 3D conv & $\sfrac{1}{6}$H$\times$$\sfrac{1}{6}$W$\times$48$\times$48\\
		$[16]$& 3$\times$3$\times$3, 3D deconv, stride 2 & $\sfrac{1}{3}$H$\times$$\sfrac{1}{3}$W$\times$48$\times$32\\
		$[17]$& 3$\times$3$\times$3, 3D conv & $\sfrac{1}{3}$H$\times$$\sfrac{1}{3}$W$\times$48$\times$32\\
		$[18]$& {\bf SGA} layer: weight matrixes from (11) &$\sfrac{1}{3}$H$\times$$\sfrac{1}{3}$W$\times$48$\times$32\\
		\multirow{4}[0]{*}{\tabincell{c}{final\\output}} &3$\times$3$\times$3, 3D to 2D conv, upsamping & H$\times$W$\times$193\\
		&{\bf LGA}, softmax, {\bf LGA}: weights from (14), (17) & H$\times$W$\times$193\\
		& regression, loss weight: 1.0 & H$\times$W$\times$1\\

		\hline
		{connection} & \multicolumn{2}{l}{concat: (4,6), (3,8), (7,11), (5,13), (12,14), (10,16); add: (1,3)} \\
		\hline
	\end{tabular}
	\label{tab:architecture}%
	\vspace*{-0.075in}
\end{table*}


{\small
\bibliographystyle{ieee}
\bibliography{arxiv}
}

\end{document}